\documentclass[journal,twoside,web]{ieeecolor}
\usepackage{generic}
\usepackage{cite}
\usepackage{amsmath,amssymb,amsfonts}
\usepackage{algorithmic}
\usepackage{graphicx}
\usepackage{algorithm,algorithmic}
\usepackage{hyperref}
\usepackage{lipsum}
\usepackage{multirow}
\usepackage{booktabs}
\hypersetup{hidelinks=true}
\usepackage{textcomp}
\def\BibTeX{{\rm B\kern-.05em{\sc i\kern-.025em b}\kern-.08em
    T\kern-.1667em\lower.7ex\hbox{E}\kern-.125emX}}
    
\begin{document}
\title{Federated Survival Analysis in Healthcare: A Multi-Model Evaluation on Cross-Institutional Heterogeneous Breast Cancer Data}
\author{Natalia Moreno-Blasco, Anusha Ihalapathirana, Pekka Siirtola, Miguel Fernandez-de-Retana
\thanks{This work was supported by the Basque Government under grant DEUSTEK6 -- Humanized Computing for Smart Sustainable and Healthier Communities and Environments (IT1901-26), and by the European Union’s Horizon Europe Research and Innovation Programme under the LATE-AYA project (Grant Agreement No. 101214326).
}
\thanks{N. Moreno-Blasco is with the Faculty of Information Technology and Electrical Engineering, University of Oulu, Oulu, Finland (e-mail: natalia.moreno@student.oulu.fi).}
\thanks{A. Ihalapathirana and P. Siirtola are with the Biomimetics and Intelligent Systems Group, University of Oulu, Oulu, Finland (e-mails: \{anusha.ihalapathirana, pekka.siirtola\}@oulu.fi).}
\thanks{M. Fernandez-de-Retana is with the Faculty of Engineering, University of Deusto, Bilbao, Spain (e-mail: m.fernandezderetana@deusto.es).}
}

\maketitle

\begin{abstract}
Survival analysis is central to clinical decision-making, yet reliable time-to-event models require large, diverse cohorts that are rarely available at a single institution, while privacy regulations restrict the centralization of patient data. Federated learning (FL) offers a privacy-preserving alternative by training shared models without exchanging raw data, but its effectiveness for survival modeling under realistic, heterogeneous conditions remains insufficiently understood. This paper presents a systematic, multi-model evaluation of federated survival analysis on a cross-institutional breast cancer cohort with naturally heterogeneous distributed clients. Three representative survival models, the Cox Proportional Hazards model, DeepSurv, and Random Survival Forest (RSF), are compared across centralized, local, and federated training, and three federated optimization strategies (FedAvg, FedProx, and FedAdam) are assessed for the gradient-based models. Results show that FL consistently outperforms local training and approaches, and occasionally exceeds, centralized performance, while RSF offers the best overall balance of discrimination, calibration, and robustness across heterogeneous clients. We further find that performance depends on the diversity of client distributions, and that FedAvg and FedProx are stronger and more stable than FedAdam. Based on these findings, we derive practical, decision-oriented guidelines mapping data, privacy, interpretability, and resource constraints to recommended model and training-paradigm choices for federated survival modeling in healthcare.
\end{abstract}

\begin{IEEEkeywords}
Breast Cancer, Federated Learning, Healthcare Data, ML Privacy, Oncology, Survival Analysis
\end{IEEEkeywords}

\section{Introduction}\label{sec:introduction}
\IEEEPARstart{T}{he} accelerating integration of artificial intelligence (AI) into clinical practice has turned data-driven systems into a central tool for tasks such as diagnosis, prognosis, treatment planning, and disease-progression forecasting~\cite{topol2019high, maslej2025artificial}. These systems can support clinicians in making more accurate and timely decisions by learning patterns from large collections of patient records~\cite{kelly2019key}. However, their deployment in healthcare is fundamentally constrained by the decentralized and privacy-sensitive nature of medical data, which is typically siloed across hospitals, laboratories, and research centers, each maintaining its own storage infrastructure, governance procedures, and ethical oversight~\cite{rieke2020future}. Even when technical interoperability is feasible, legal and ethical frameworks such as the General Data Protection Regulation (GDPR)~\cite{gdpr2016} in Europe and the Health Insurance Portability and Accountability Act (HIPAA)~\cite{hipaa} in the United States restrict the centralization of patient information.

This situation creates a tension between data accessibility and data privacy. On the one hand, robust machine learning (ML) models require large, diverse, and representative datasets to achieve generalizable predictions; on the other hand, the sensitivity of clinical data makes unrestricted aggregation neither acceptable nor viable. As a consequence, many studies rely on single-institution datasets, producing models that perform well in one clinical setting but fail to generalize across others, a phenomenon commonly referred to as dataset bias or domain shift~\cite{kelly2019key}. Beyond predictive performance, models trained on homogeneous data can also reflect the demographic or procedural biases of the institutions where the data were collected, raising important ethical and fairness concerns~\cite{chen2021ethical}.

Federated learning (FL), introduced by McMahan et al.~\cite{mcmahan2017communication}, has emerged as a promising paradigm to address these challenges. Instead of pooling data in a single location, FL enables multiple institutions to collaboratively train a shared model by exchanging only model parameters or updates, while keeping the underlying patient data local and private. This approach preserves confidentiality, promotes cross-institutional cooperation, and improves generalization by exposing the model to a broader range of data distributions~\cite{rieke2020future}. FL has already shown promise in healthcare applications such as multi-institutional prediction of clinical outcomes in COVID-19 patients~\cite{dayan2021federated} and privacy-preserving brain tumour segmentation~\cite{li2019privacy}. Nevertheless, FL in healthcare remains an emerging field~\cite{kairouz2021advances}: practical deployments must contend with communication overhead, statistical heterogeneity across clients (non-independent and identically distributed, or non-\textit{iid}, data)~\cite{Li2020FedAvgNonIID}, and the fact that sharing model updates does not provide absolute privacy, as gradient-leakage attacks can partially reconstruct private inputs~\cite{geiping2020inverting, zhu2019deep, fernandez2025differential}.

A clinically central yet comparatively underexplored application of FL is survival analysis, the branch of statistics concerned with modeling the time until an event of interest, such as disease progression, relapse, or death, occurs~\cite{clark2003survival, kleinbaum1996survival}. Unlike standard regression or classification, survival models must explicitly account for censoring, that is, individuals for whom the event has not been observed within the study period. Survival analysis plays a decisive role in oncology, where breast cancer is the most frequently diagnosed cancer worldwide~\cite{sung2021global} and where reliable time-to-event estimates inform prognosis and treatment decisions. Because such models benefit from large and diverse cohorts that are rarely available at a single institution, survival analysis is a natural candidate for federated settings; at the same time, censoring, unbalanced event rates, and inter-institutional differences make the federated formulation methodologically challenging.

The combination of FL and survival analysis is a recent but rapidly growing research direction. FedSurF++~\cite{archetti2023scaling} extends Random Survival Forests (RSF) to the federated setting by aggregating locally trained survival trees in a single communication round. Andreux et al.~\cite{andreux2020federated} showed that naively federating the Cox Proportional Hazards (CoxPH) model yields a stratified Cox formulation that degrades under heterogeneity, and proposed a discrete-time reformulation with a separable loss to enable effective federated training. More recently, FedScore-Surv~\cite{li2025developing} developed privacy-preserving federated time-to-event scores across institutions, and FedPseudo~\cite{rahman2022fedpseudo} introduced a pseudo-value-based deep learning framework for federated survival modeling. These contributions complement a broader body of FL-in-healthcare work spanning intensive-care mortality prediction~\cite{mondrejevski2022flicu}, multi-modal COVID-19 diagnosis~\cite{qayyum2022collaborative}, anomaly detection in the Internet of Medical Things (IoMT)~\cite{gupta2021hierarchical}, and cross-federation knowledge distillation~\cite{chen2023metafed}.

Despite this progress, most existing studies either focus on a single model family or propose a new algorithm evaluated in isolation. Few works systematically compare different model families across training paradigms under controlled data heterogeneity, and ensemble methods such as RSF remain largely unexplored in federated settings beyond FedSurF++. Crucially, there is a lack of practical, decision-oriented guidance on \emph{which} survival model and \emph{which} training paradigm to choose under given data characteristics, privacy constraints, and computational budgets. This gap motivates a regime-based, comparative evaluation rather than the pursuit of a single best-performing model.

In this work\footnote{Code publicly available at: \href{https://github.com/nataliamorenob/Survival-Models-in-Federated-Healthcare-Settings}{https://github.com/nataliamorenob/Survival-Models-in-Federated-Healthcare-Settings}}, we present a systematic, multi-model evaluation of federated survival analysis on cross-institutional, heterogeneous breast cancer data. We compare three representative survival models, namely the statistical Cox Proportional Hazards (CoxPH) model, the deep-learning-based DeepSurv model, and the tree-ensemble Random Survival Forest (RSF) model, across three training paradigms (centralized, local, and federated), using the Fed-TCGA-BRCA dataset from the FLamby benchmark suite~\cite{ogier2022flamby}. The main contributions of this paper are as follows:
\begin{itemize}
    \item A unified empirical comparison of CoxPH, DeepSurv, and RSF under local, federated, and centralized training on a real, multi-institutional breast cancer cohort with naturally heterogeneous client distributions.
    \item An analysis of how data heterogeneity, expressed through the number and composition of participating clients, affects both discrimination and calibration of federated survival models.
    \item An evaluation of federated optimization strategies (FedAvg, FedProx, and FedAdam) for the two gradient-based survival models, assessing their robustness under heterogeneous client distributions.
    \item A set of practical, decision-oriented guidelines that map data, privacy, interpretability, and resource constraints to recommended model and training-paradigm choices.
\end{itemize}

The remainder of this paper is organized as follows. Section~\ref{sec:methods} describes the dataset, the survival models, the learning paradigms and federated optimization strategies, and the experimental setup. Section~\ref{sec:results} reports the evaluation metrics and experimental results, Section~\ref{sec:discussion} discusses the main findings and derives practical guidelines, and the final section concludes the paper and outlines future research directions.

\section{Methods}\label{sec:methods}

This section describes the methodological framework adopted in this study. First, the Fed-TCGA-BRCA dataset is presented, including a characterization of its cross-institutional heterogeneity. Then, the three survival models are introduced together with their suitability for federated time-to-event modeling. Finally, the learning paradigms, the federated optimization strategies, and the experimental setup are detailed.

\subsection{Dataset and Preprocessing}
\label{sec:dataset}

The experiments are conducted on the Fed-TCGA-BRCA dataset, obtained from the FLamby benchmark suite~\cite{ogier2022flamby}, which is specifically designed for cross-silo federated learning in realistic healthcare settings. The underlying data originate from The Cancer Genome Atlas Program (TCGA)~\cite{weinstein2013cancer}, one of the largest publicly available cancer genomics resources, accessible through the Genomic Data Commons (GDC) data portal~\cite{heath2021nci}. The cohort focuses on breast invasive carcinoma (BRCA) and integrates the clinical, pathological, and molecular profiles of more than one thousand patients collected across multiple institutions worldwide. For the FLamby benchmark, only the clinical tabular subset of TCGA-BRCA is used, as it provides structured, interpretable variables suitable for tabular ML and allows modeling of the time from diagnosis or treatment initiation until death or last follow-up.

Each patient is represented by $39$ features grouped into demographic (e.g., age, ethnicity, and race indicators), pathological (e.g., tumor, node, metastasis, and overall stage variables), diagnostic and coding (ICD-10 and morphology indicators), clinical history and treatment (prior malignancy and therapy indicators), and simplified tumor-stage variables. The outcome is described by an event indicator $E$ (i.e., $1$ for an observed death, $0$ for a right-censored observation) and a survival time $T$ measured in days. The dataset is partitioned into training and test sets according to the original FLamby design. The training set is further split into local training and validation sets within each client, as described in Section~\ref{sec:experimental_setup}.

Importantly, a defining property of this dataset is its natural cross-institutional \textit{heterogeneity}. Table~\ref{tab:dataset_summary} summarizes the cohort across the six original centers, reporting the number of patients, observed events (deaths), censored cases, the range of observed follow-up times, and the event rate. The dataset comprises $1{,}088$ patients with $151$ recorded events and $937$ censored observations, yielding an overall event rate of approximately $13.9\%$. The centers differ markedly in size (from $51$ to $311$ patients), event rate (from $5.6\%$ to $19.9\%$), and maximum observed follow-up (from $1{,}900$ to $8{,}605$ days), and they also exhibit differing age distributions and survival profiles. This combination of variable sample sizes, censoring levels, and follow-up windows reproduces the non-\textit{iid} conditions encountered in real multi-institutional studies and makes the dataset well suited for evaluating federated survival models under realistic heterogeneity.

\begin{table*}[t]
\caption{Summary of the Fed-TCGA-BRCA Dataset by Center (Combined Training and Test Data)}
\label{tab:dataset_summary}
\centering
\footnotesize
\setlength{\tabcolsep}{8pt}
\renewcommand{\arraystretch}{1.15}
\begin{tabular}{lcccccc}
\hline
\textbf{Center} & \textbf{Patients ($N$)} & \textbf{Events ($N_{\text{events}}$)} & \textbf{Censored ($N_{\text{censored}}$)} & \textbf{Min Time (days)} & \textbf{Max Time (days)} & \textbf{Event Rate (\%)} \\
\hline
0 & 311 & 59 & 252 & 64.0 & 4047.0 & 19.0 \\
1 & 196 & 39 & 157 & 78.0 & 8605.0 & 19.9 \\
2 & 206 & 22 & 184 & 1.0 & 4354.0 & 10.7 \\
3 & 162 & 19 & 143 & 21.0 & 5062.0 & 11.7 \\
4 & 162 & 9  & 153 & 0.0 & 3409.0 & 5.6 \\
5 & 51  & 3  & 48  & 0.0 & 1900.0 & 5.9 \\
\hline
\textbf{Total} & \textbf{1088} & \textbf{151} & \textbf{937} & \textbf{0.0} & \textbf{8605.0} & \textbf{13.9} \\
\hline
\end{tabular}
\vspace{1mm}

\footnotesize
\textit{Note:} Experiments use up to five clients, mapped to centers $\{0,\ldots,4\}$, as described in Section~\ref{sec:experimental_setup}. Client 5 was excluded due to limited events.
\end{table*}

Prior to training, the data of each client are split using a stratified, event-aware strategy that preserves the event-to-censoring ratio and avoids data leakage. The held-out test partition is the one provided by FLamby; the remaining data are further divided into training and validation sets, with the validation proportion adapted to the local number of observed events to guarantee sufficient event representation:
\begin{equation}
\text{val\_size} =
\begin{cases}
0.30 & \text{if } n_{\text{events}} \ge 20 \\
0.25 & \text{if } 10 \le n_{\text{events}} < 20 \\
0.20 & \text{if } n_{\text{events}} < 10
\end{cases}
\end{equation}
with the additional constraint that each validation set contains at least two uncensored events. Continuous features are standardized using \textit{per-client} $z$-score normalization, with the mean and standard deviation computed exclusively on the training data and subsequently applied to the validation and test partitions to avoid leakage; the remaining binary features are left unscaled.

\subsection{Survival Models}
\label{sec:models}

Survival analysis deals with time-to-event data in the presence of censoring. For each individual $i$, let $T_i$ denote the (possibly unobserved) event time and $C_i$ the censoring time; the observed follow-up time and event indicator are:
\begin{equation}
Y_i = \min(T_i, C_i), \qquad \delta_i = \mathbb{I}(T_i \le C_i)
\end{equation}
so that $\delta_i = 1$ when the event is observed and $\delta_i = 0$ when the observation is right-censored. Together with a covariate vector $\mathbf{x}_i \in \mathbb{R}^p$, the observed data form a collection of triplets $(\mathbf{x}_i, Y_i, \delta_i)$. The two quantities of interest are the survival function $S(t) = \mathbb{P}(T > t)$ and the hazard function: 
\begin{equation}
h(t) = \lim_{\Delta t \to 0} \dfrac{\mathbb{P}(t \le T < t + \Delta t \mid T \ge t)}{\Delta t}
\end{equation}
which represents the instantaneous risk of experiencing the event at time $t$ given survival up to that time. Survival models aim to estimate these functions as a function of the covariates $\mathbf{x}$, enabling predictions of individual survival probabilities and risk scores.

To compare complementary modeling philosophies, three survival models with native support for censored data are evaluated: the statistical CoxPH model, the deep-learning-based DeepSurv model, and the tree-ensemble RSF model. Each model is trained and evaluated under the centralized, local, and federated paradigms described in Section~\ref{sec:paradigms}.

\subsubsection{Cox Proportional Hazards (CoxPH)}
The CoxPH model~\cite{cox1972regression} expresses the hazard of an individual as the product of an unspecified baseline hazard $h_0(t)$ and an exponential function of the covariates:
\begin{equation}
h(t \mid \mathbf{x}_i) = h_0(t)\,\exp\!\left(\boldsymbol{\beta}^{\top} \mathbf{x}_i\right)
\end{equation}
under the \textit{proportional-hazards} assumption that the hazard ratio between any two individuals is constant over time. Being semi-parametric, the model estimates the coefficients $\boldsymbol{\beta}$ through partial-likelihood maximization without specifying $h_0(t)$. Its interpretability, simplicity, and long history make CoxPH a standard reference in clinical survival analysis~\cite{kleinbaum1996survival}. A known limitation in distributed settings, however, is its reliance on global risk sets: the partial likelihood couples individuals across the whole cohort, information that is not locally available when data cannot be shared~\cite{andreux2020federated}. In this work, CoxPH serves as a classical statistical baseline against which more flexible models are compared.

\subsubsection{DeepSurv}
DeepSurv~\cite{katzman2018deepsurv} is a deep-learning extension of CoxPH in which the linear risk term $\boldsymbol{\beta}^{\top}\mathbf{x}$ is replaced by a neural network $g(\mathbf{x};\boldsymbol{\theta})$, allowing nonlinear covariate effects to be modeled. The network consists of an input layer, hidden layers with nonlinear activations, and a single output that predicts a log-risk score; and is trained by minimizing the negative Cox partial log-likelihood using gradient-based optimization. DeepSurv is a widely adopted neural survival baseline, and several related deep models, such as DeepHit~\cite{lee2018deephit} or VAECox~\cite{kim2020improved} relax the proportional-hazards assumption further. Because DeepSurv inherits the same partial-likelihood loss as CoxPH, the Cox loss must be approximated locally in the federated setting, since the exact global risk sets cannot be reconstructed without exchanging survival-time information between clients.

\subsubsection{Random Survival Forests (RSF)}
RSF~\cite{ishwaran2008random} extends Random Forests~\cite{breiman2001random} to survival data by growing an ensemble of survival trees, each trained on a bootstrap sample and split using survival-specific criteria, and aggregating their cumulative-hazard estimates. RSF captures nonlinear relationships and covariate interactions without parametric assumptions, and has demonstrated strong predictive performance on heterogeneous clinical datasets~\cite{liao2024random}. In the federated setting, RSF follows the FedSurF++ approach~\cite{archetti2023scaling}, which operates on fully trained trees rather than gradients. In a local-training step, each client $k$ fits a forest on its local data and evaluates the quality of its trees on a local validation set using the concordance index, while sending only metadata (the number of trees $T_k$ and samples $N_k$) to the server. The server then performs a tree-assignment step, distributing the $T$ global slots among clients with probability proportional to $N_k$; finally, in a tree-sampling step, each client samples its allocated trees without replacement, with selection probability proportional to validation performance, and sends them to the server, which aggregates them into a global federated forest. Because aggregation operates directly on trees, the federated procedure requires only a single communication round, substantially reducing communication overhead relative to iterative gradient-based methods.

\subsection{Learning Paradigms}
\label{sec:paradigms}

To assess the effect of data decentralization on survival modeling, three learning paradigms that differ in how data are accessed during training are considered.

\paragraph{Centralized learning} All client data are pooled into a single set before training, and the model is trained as if originating from one source. After training, the model is evaluated on each client's local test set. This paradigm assumes that data sharing is possible and serves as a best-case reference.

\paragraph{Local learning} Each client trains an independent model on its own data, with no sharing of data, parameters, or updates. This fully privacy-preserving but non-collaborative setting provides a lower-reference baseline whose performance is expected to suffer under data scarcity and heterogeneity.

\paragraph{Federated learning} The training data remain at each institution, and a shared global model is obtained through repeated communication between the clients and a central server. Assuming $K$ clients, where client $k$ holds a local dataset $\mathcal{D}_k$ of size $n_k$ and $N = \sum_{k=1}^{K} n_k$, the federated objective minimizes the data-weighted average of the local empirical risks:
\begin{align}
f(w) &= \sum_{k=1}^{K} \frac{n_k}{N}\, F_k(w) \\
F_k(w) &= \frac{1}{n_k} \sum_{(x_i,y_i)\in\mathcal{D}_k} \ell(w; x_i, y_i)
\end{align}
where $w$ denotes the model parameters and $\ell$ the per-sample loss. At each round, the server broadcasts the current global model, each selected client performs local optimization steps, and the resulting updates are aggregated into a new global model; the process repeats until convergence. The final global model is evaluated on each client's local test set.

\subsection{Federated Optimization Strategies}
\label{sec:strategies}

For the two gradient-based models (CoxPH and DeepSurv), three federated aggregation and optimization strategies are evaluated. The RSF model relies on tree aggregation rather than gradient-based optimization and is therefore incompatible with these strategies.

\paragraph{FedAvg} Federated Averaging~\cite{mcmahan2017communication} is the standard FL baseline. After a few local stochastic-gradient-descent steps, the server computes the data-weighted average of the received parameters:
\begin{equation}
w^{(t+1)} = \sum_{k \in S_t} \frac{n_k}{N}\, w_k^{(t+1)}
\end{equation}
where $S_t$ is the set of clients selected at round $t$. FedAvg is communication-efficient but assumes that local updates do not diverge excessively, which may not hold under strong non-\textit{iid} conditions.

\paragraph{FedProx} FedProx~\cite{li2020federated} extends FedAvg by adding a proximal term to each local objective:
\begin{equation}
\min_{w}\; F_k(w) + \frac{\mu}{2}\, \lVert w - w^{(t)} \rVert^2
\end{equation}
where $w^{(t)}$ is the current global model and $\mu$ controls the penalty on local drift. The proximal term limits divergence between local and global models, which is designed to improve robustness under heterogeneous client distributions and variable local computation.

\paragraph{FedAdam} FedAdam~\cite{reddi2020adaptive} applies adaptive moment estimation on the server side. Rather than averaging updates directly, the server maintains first- and second-moment estimates of the aggregated update $\Delta_t$:
\begin{align}
m_t &= \beta_1 m_{t-1} + (1 - \beta_1)\, \Delta_t \\
v_t &= \beta_2 v_{t-1} + (1 - \beta_2)\, \Delta_t^2
\end{align}
where $\beta_1$ and $\beta_2$ are exponential decay rates, and updates the global model using an Adam-style rule \cite{kingma2014adam}. This adaptive server optimization is intended to improve convergence stability for non-convex objectives.

Federated orchestration and communication are implemented using the Flower framework~\cite{beutel2020flower}, and the number of communication rounds is derived from the data-driven formulation provided by FLamby (Section~\ref{sec:experimental_setup}).

\subsection{Experimental Setup}
\label{sec:experimental_setup}

To study the effect of data heterogeneity, experiments are performed with three client configurations of five, four, and three clients, mapped consistently to centers $\{0,1,2,3,4\}$, $\{0,1,2,3\}$, and $\{0,1,2\}$, respectively. Reducing the number of clients excludes specific data distributions rather than redistributing data, so each remaining client retains its original distribution. To account for stochasticity in training, each experiment is repeated ten times with different random seeds, and all reported metrics are given as the mean and standard deviation over these runs. 

For a fair comparison between centralized and federated training, the maximum number of federated rounds is not chosen arbitrarily but computed so that the total number of local update steps matches the centralized training budget. Following FLamby~\cite{ogier2022flamby}, given $n^{P}_{\text{epochs}}$ centralized epochs, $n_T$ total training samples, $K$ clients, batch size $B$, and $E$ local update steps:
\begin{equation}
T_{\max} = n^{P}_{\text{epochs}} \left\lfloor \frac{n_T}{K \cdot B \cdot E} \right\rfloor
\end{equation}
For RSF, a separate design is used because of its tree-based nature: a first experiment selects the number of local trees (varied among $20$, $50$, $100$, and $200$) while keeping the number of aggregated global trees fixed, and a second experiment evaluates the chosen configuration across paradigms and client counts. After this analysis, the number of trees is fixed at $100$, which balances performance and computational cost.

Importantly, the held-out test partition is never used during training or validation. In federated experiments, evaluation is performed after each communication round, and the final results are reported at the last round, averaged across runs. For time-dependent metrics, a client-specific grid of $100$ equally spaced time points is defined over the common follow-up interval shared by the client's training and test sets. Let $\{t^{\mathrm{tr}}_{(1)} \le \cdots \le t^{\mathrm{tr}}_{(n)}\}$ and $\{t^{\mathrm{te}}_{(1)} \le \cdots \le t^{\mathrm{te}}_{(m)}\}$ denote the ordered observed times in the training and test sets; the grid bounds are:
\begin{equation}
t_{\min} = \max\!\left(t^{\mathrm{tr}}_{(2)}, t^{\mathrm{te}}_{(2)}\right),
\;
t_{\max} = \min\!\left(t^{\mathrm{tr}}_{(n-1)}, t^{\mathrm{te}}_{(m-1)}\right)
\end{equation}
and the evaluation grid is $\mathcal{T} = \{\, t_{\min} + \tfrac{k}{99}(t_{\max} - t_{\min})\}_{k=0}^{99}$. Restricting evaluation to this common interior interval avoids boundary regions where the risk set is small and estimates based on censored times become unstable.

Finally, all three models are evaluated under the centralized, local, and federated paradigms for the five-, four-, and three-client configurations. For the gradient-based models, the federated setting is additionally evaluated with FedAvg, FedProx, and FedAdam; for clarity, the main results report FedAvg, which is consistently strong and stable, while the comparison across strategies is analyzed separately. The performance of all models is assessed using the discrimination and calibration metrics described next.

\section{Results}\label{sec:results}
This section presents the experimental results obtained for the evaluated survival analysis models and FL configurations. First, the evaluation metrics used in the study are described. Subsequently, the individual performance of the CoxPH, DeepSurv, and RSF models is analyzed across the different training paradigms to further investigate the impact of distributed learning settings on survival prediction performance, followed by a cross-model comparison. Finally, the effect of different federated optimization strategies, namely FedAvg, FedAdam, and FedProx, is analyzed to further assess the robustness of federated optimization under heterogeneous client distributions in the evaluated experimental setting.

\subsection{Evaluation Metrics}

This section describes the evaluation metrics used to assess the performance of the survival analysis models across different training paradigms and client configurations. The selected metrics include both \textit{discrimination} and \textit{calibration} measures, which are essential for evaluating the predictive performance of survival models in a comprehensive manner. Discrimination metrics, such as the Concordance Index (C-Index) and the time-dependent Area Under the Curve (AUC), evaluate the model's ability to correctly rank individuals according to their risk of experiencing the event. On the other hand, calibration metrics, such as the Integrated Brier Score (IBS), assess the accuracy of the predicted probabilities of survival over time. By using a combination of these complementary metrics, it is possible to obtain a more complete understanding of the strengths and weaknesses of each model, which can provide a broader and more reliable evaluation of model performance, as each metric captures different aspects of predictive behavior.

\subsubsection{Concordance Index (C-Index)}
    Introduced by Harrell et al.~\cite{Harrell1982}, it measures the ability of a survival model to rank individuals correctly according to their risk of experiencing the event while accounting for censoring. It is calculated as the proportion of all pairs of comparable individuals for which the survival model assigns a higher risk to the individual who experiences the event first. A C-Index of 0.5 is equivalent to random ranking, while a C-Index of 1.0 is equivalent to perfect discrimination. The C-Index is defined as:
    \begin{equation}
    \text{C-Index} =
    \frac{
    \sum_{i}\sum_{j}
    \mathbb{I}(r_i > r_j)
    \mathbb{I}(T_i < T_j)
    \mathbb{I}(\delta_i = 1)
    }{
    \sum_{i}\sum_{j}
    \mathbb{I}(T_i < T_j)
    \mathbb{I}(\delta_i = 1)
    }
    \label{eq:cindex}
    \end{equation}
    where $r_i$ and $r_j$ represent the predicted risk scores for individuals $i$ and $j$ respectively, $T_i$ and $T_j$ are the observed event or censoring times, $\delta_i$ is the event indicator (1 if the event occurred, 0 if censored), and $\mathbb{I}(\cdot)$ is the indicator function that equals 1 if the condition is true and 0 otherwise.

\subsubsection{Time-Dependent Area Under the Curve (AUC)} 
    Based on the cumulative/dynamic ROC formulation proposed by Heagerty et al.~\cite{Heagerty2000}, it evaluates the model’s capacity to distinguish between individuals who have experienced the event before a certain time and those who have not experienced the event, accounting for censoring. In this work, the AUC is reported as the time-integrated cumulative dynamic AUC evaluated on the time grid of the evaluation times. The time-dependent AUC at time \( t \) is defined as:
    \begin{equation}
    \text{AUC}(t) =
    \mathbb{P}
    \left(
    r_i(t) > r_j(t)
    \;\middle|\;
    T_i \le t,\;
    \delta_i = 1,\;
    T_j > t
    \right)
    \end{equation}
    where $i$ and $j$ index two distinct individuals, $r_i(t)$ represents the time-dependent predicted risk score for individual $i$ at time $t$, and $T_i$ is the observed event or censoring time for individual $i$. The condition that $T_i \le t$ for individual $i$ implies that individual $i$ experienced the event before or at time $t$, and that $T_j > t$ for individual $j$ implies that individual $j$ is event-free at time $t$. The time-integrated AUC is then calculated by integrating the time-dependent AUC over the specified time interval:
    \begin{equation}
    \text{AUC}_{\text{int}} =
    \frac{1}{t_{\max} - t_{\min}}
    \int_{t_{\min}}^{t_{\max}}
    \text{AUC}(t)\,dt
    \end{equation}
    where $t_{\min}$ and $t_{\max}$ specify the window of evaluation.
    The integrated AUC is a summary of the average discriminative ability of the model over the interval from $t_{\min}$ to $t_{\max}$.

\subsubsection{Integrated Brier Score (IBS)}
    Defined by Graf et al.~\cite{Graf1999}, it assesses the accuracy of the predicted probabilities of survival over time. It calculates the squared error of the predicted probabilities of survival and the observed outcome and integrates this error over the evaluation time interval. A lower value of IBS is an indicator of a more accurate and better-calibrated survival model. The Brier score at time $t$ is given by:
    \begin{equation}
    \text{BS}(t) =
    \frac{1}{N}
    \sum_{i=1}^{n}
    \left(
    \mathbb{I}(T_i > t) - \hat{S}_i(t)
    \right)^2
    \end{equation}
    where $N$ is the total number of individuals, $T_i$ is the observed event or the censoring time for individual $i$, $\hat{S}_i(t)$ is the predicted survival probability for individual $i$ at time $t$, and $\mathbb{I}(T_i > t)$ is the indicator function, equal to 1 if individual $i$ is event-free at time $t$ and 0 otherwise. The Integrated Brier Score (IBS) is then given by:
    \begin{equation}
    \text{IBS} =
    \frac{1}{t_{\max} - t_{\min}}
    \int_{t_{\min}}^{t_{\max}}
    \text{BS}(t)\,dt
    \end{equation}
    where $t_{\min}$ and $t_{\max}$ define the evaluation time interval. The IBS represents the average squared prediction error over $[t_{\min}, t_{\max}]$, providing a scalar summary of overall predictive accuracy. Lower values indicate better predictive performance and calibration.

\subsection{Model Performance Under Different Training Paradigms}

This section presents a detailed analysis of the performance of the CoxPH, DeepSurv, and RSF models across the \textit{local}, \textit{federated}, and \textit{centralized} training paradigms for different client configurations. The analysis aims to evaluate the effect of distributed learning settings on survival prediction performance, while also examining client-level variability under heterogeneous data distributions. Finally, a client-level cross-model comparison is conducted under the federated learning setting to analyze the relative predictive performance of the evaluated survival models across different clients and evaluation metrics. The complete per-client results underlying these analyses are reported in Appendix~\ref{app:client_tables}.

\subsubsection{CoxPH}
\label{sec:coxph_results}
The average performance of the CoxPH models using centralized, federated, and local training methods across different numbers of clients is presented in Table~\ref{tab:coxph_mean}. The best discrimination performance can be obtained when using the centralized training method in terms of both C-Index and AUC scores. When five clients are present, the performance of centralized training is measured at 0.732 on the C-Index, in contrast to federated and local training performances of 0.611 and 0.598, respectively. A similar trend in terms of AUC was observed, where centralized training obtained a value of 0.697 compared to federated (0.600) and local (0.606) training.

Regarding calibration performance, federated training achieved the lowest IBS values across all client configurations. For instance, with five clients, federated training obtained an IBS of 0.155, compared to 0.167 for local training and 0.290 for centralized training. As the number of clients decreased from five to three, there were minor changes in results across all training paradigms. In general, it was observed that C-Index and AUC values remained relatively stable, whereas IBS values increased for local and federated training.

\begin{table}[t]
\caption{Mean CoxPH Performance Across Training Paradigms and Number of Clients}
\label{tab:coxph_mean}
\centering
\footnotesize
\setlength{\tabcolsep}{6pt}
\renewcommand{\arraystretch}{1.15}

\begin{tabular}{lcccc}
\hline
\textbf{Training Paradigm} & \textbf{Clients} & \textbf{C-Index} ($\uparrow$) & \textbf{AUC} ($\uparrow$) & \textbf{IBS} ($\downarrow$) \\
\hline

\multirow{3}{*}{Local}
    & 5 & 0.598 & 0.606$^{*}$ & 0.167 \\
    & 4 & 0.566 & 0.564        & 0.197 \\
    & 3 & 0.570 & 0.554        & 0.227 \\
\hline

\multirow{3}{*}{Federated (FedAvg)}
    & 5 & 0.611        & 0.600$^{*}$ & 0.155 \\
    & 4 & 0.594        & 0.581$^{*}$ & 0.183 \\
    & 3 & 0.609$^{*}$ & 0.573        & 0.213 \\
\hline

\multirow{3}{*}{Centralized}
    & 5 & 0.732 & 0.697 & 0.290 \\
    & 4 & 0.702 & 0.659 & 0.288 \\
    & 3 & 0.699 & 0.658 & 0.368 \\
\hline

\end{tabular}

\vspace{1mm}

\footnotesize
\textit{Note:} $^{*}$Values do not exhibit statistically significant differences with Centralized according to Dunn's post-hoc test ($p > 0.05$).
\end{table}

\begin{figure*}[t]
\centering
\includegraphics[width=0.9\textwidth]{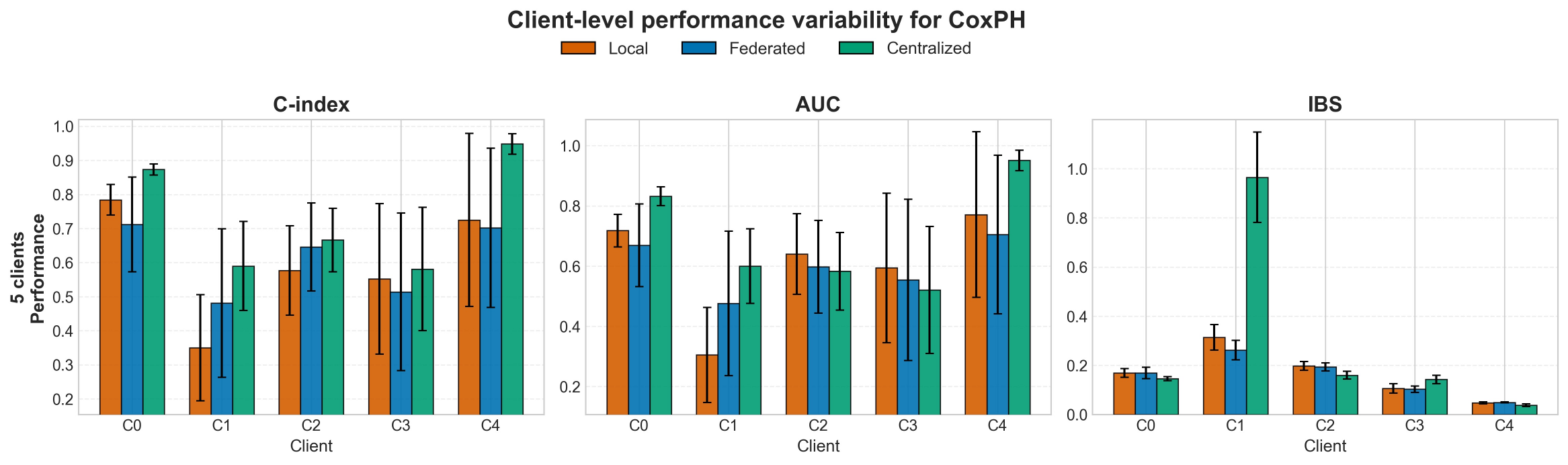}
\caption{Client-level performance variability across training paradigms for CoxPH under the five-client configuration. Results are reported as mean $\pm$ standard deviation across 10 independent runs for each client: (a) C-Index, (b) AUC, and (c) IBS.}
\label{fig:coxph_clients}
\end{figure*}

The variation in performance of the CoxPH algorithm from the point of view of clients is illustrated in Fig.~\ref{fig:coxph_clients} under the five-client configuration. This setup was selected as it represents the scenario with the highest number of participating clients. Variation between performance can be seen among all the metrics under all the training paradigms. Clients C0 and C4 achieved the highest C-Index and AUC values, respectively, for most of the training paradigms; whereas it was observed that client C1 achieved a lower performance. Variability across runs, represented by the error bars, was higher for certain clients, particularly under local and federated training. Lastly, across several clients, federated training tended to outperform local training on discrimination metrics, though this advantage was not consistent for all clients and metrics.

\subsubsection{DeepSurv}
\label{sec:deepsurv_results}
Table~\ref{tab:deepsurv_mean} summarizes the performance of the DeepSurv model under local, federated, and centralized training paradigms across different client configurations. Federated training achieved the highest discrimination performance for most client configurations, obtaining the highest C-Index and AUC values. When considering the five-client configuration, federated training reached a C-Index of 0.727, compared to 0.707 for centralized training and 0.618 for local training. A similar trend was observed for AUC, where federated training achieved a value of 0.702, outperforming centralized and local learning, whose scores were 0.684 and 0.620, respectively.

In terms of IBS, federated training also obtained the lowest values across all client configurations. For instance, with five clients, federated training achieved an IBS of 0.148, compared to 0.159 for local training and 0.269 for centralized training. As the number of clients decreased from five to three, minor differences in performance were observed across all paradigms. Overall, federated learning offered strong discrimination performance while consistently achieving the lowest IBS values across all client configurations.

\begin{table}[t]
\caption{Mean DeepSurv Performance Across Training Paradigms and Number of Clients}
\label{tab:deepsurv_mean}
\centering
\footnotesize
\setlength{\tabcolsep}{6pt}
\renewcommand{\arraystretch}{1.15}

\begin{tabular}{lcccc}
\hline
\textbf{Training Paradigm} & \textbf{Clients} & \textbf{C-Index} ($\uparrow$) & \textbf{AUC} ($\uparrow$) & \textbf{IBS} ($\downarrow$) \\
\hline

\multirow{3}{*}{Local}
    & 5 & 0.618$^{*}$ & 0.620 & 0.159 \\
    & 4 & 0.578$^{*}$ & 0.573 & 0.187$^{*}$ \\
    & 3 & 0.601$^{*}$ & 0.595 & 0.215$^{*}$ \\
\hline

\multirow{3}{*}{Federated (FedAvg)}
    & 5 & 0.727$^{*}$ & 0.702 & 0.148 \\
    & 4 & 0.671$^{*}$ & 0.636 & 0.174 \\
    & 3 & 0.724$^{*}$ & 0.667 & 0.198$^{*}$ \\
\hline

\multirow{3}{*}{Centralized}
    & 5 & 0.707 & 0.684 & 0.269 \\
    & 4 & 0.677 & 0.651 & 0.288 \\
    & 3 & 0.650 & 0.616 & 0.347 \\
\hline

\end{tabular}

\vspace{1mm}

\footnotesize
\textit{Note:} $^{*}$Values do not exhibit statistically significant differences with Centralized according to Dunn's post-hoc test ($p > 0.05$).
\end{table}

\begin{figure*}[t]
\centering
\includegraphics[width=0.9\textwidth]{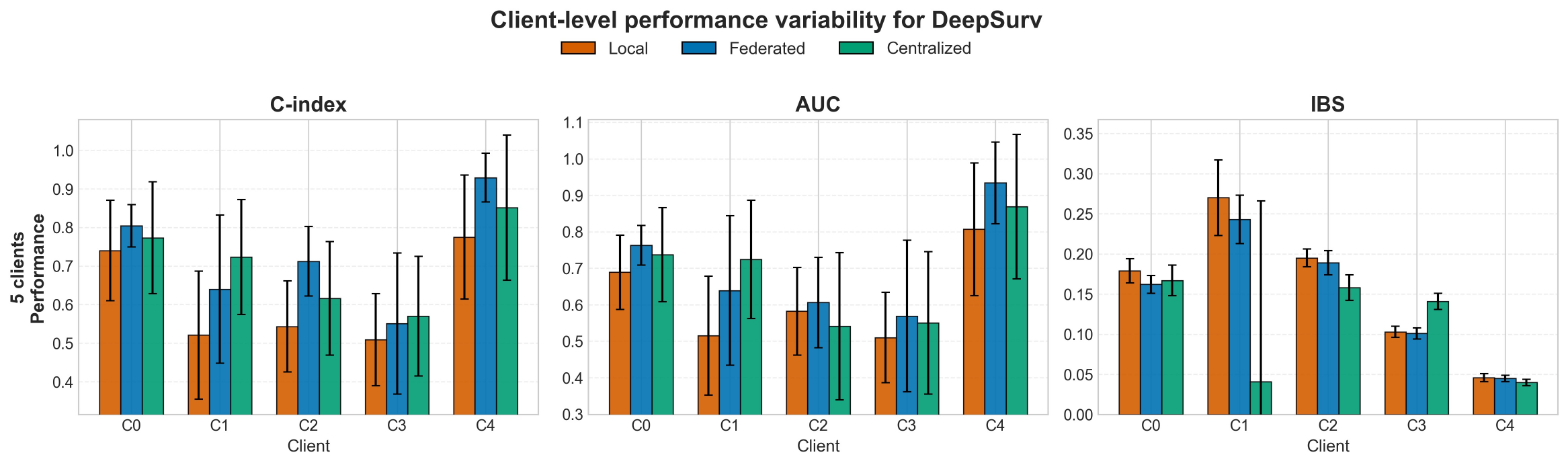}
\caption{Client-level performance variability across training paradigms for DeepSurv under the five-client configuration. Results are reported as mean $\pm$ standard deviation across 10 independent runs for each client: (a) C-Index, (b) AUC, and (c) IBS.}
\label{fig:deepsurv_clients}
\end{figure*}

Fig.~\ref{fig:deepsurv_clients} illustrates the client-level results obtained by the DeepSurv model for the five-client configuration. Differences in performance among clients were observed for all evaluated metrics and training paradigms. Client C4 obtained the highest C-Index and AUC values, whereas lower discrimination performance was observed for client C3. On the other hand, for the IBS metric, higher values were observed for client C1 under local and federated training. Variability across runs, represented by the error bars, was more pronounced for clients C1 and C4 across the discrimination metrics. Lastly, federated training achieved higher discrimination performance than local training for several clients, particularly for clients C1, C2, and C4 in terms of C-Index and AUC.

\subsubsection{Random Survival Forest (RSF)}
\label{sec:results_rsf}
The performance obtained by the RSF model under local, federated, and centralized training paradigms is reported in Table~\ref{tab:rsf_mean}. For consistency, the number of trees was fixed at 100 across all RSF configurations after preliminary experiments on different numbers of trees (i.e., 50, 100, and 200) showed that this setting provided a good balance between performance and computational efficiency. Centralized training achieved the highest C-Index values across all client configurations, reaching 0.746 for the five-client configuration, compared to 0.724 for federated training and 0.687 for local training. A similar trend was observed for AUC, where centralized training obtained the highest values for all configurations.

For the IBS metric, federated training consistently achieved the lowest values across all client configurations. For the five-client configuration, federated learning achieved an IBS of 0.138, compared to 0.152 for local learning and 0.245 for centralized learning. As the number of clients decreased from five to three, variations in performance were observed across all paradigms. In general, IBS values increased as the number of clients decreased, whereas C-Index and AUC showed moderate fluctuations.

\begin{table}[t]
\caption{Mean RSF Performance Across Training Paradigms and Number of Clients (100-Tree Configuration)}
\label{tab:rsf_mean}
\centering
\footnotesize
\setlength{\tabcolsep}{6pt}
\renewcommand{\arraystretch}{1.15}

\begin{tabular}{lcccc}
\hline
\textbf{Training Paradigm} & \textbf{Clients} & \textbf{C-Index} ($\uparrow$) & \textbf{AUC} ($\uparrow$) & \textbf{IBS} ($\downarrow$) \\
\hline

\multirow{3}{*}{Local}
    & 5 & 0.687 & 0.665 & 0.152 \\
    & 4 & 0.615 & 0.594 & 0.180$^{*}$ \\
    & 3 & 0.617 & 0.588 & 0.207$^{*}$ \\
\hline

\multirow{3}{*}{Federated}
    & 5 & 0.724$^{*}$ & 0.678$^{*}$ & 0.138 \\
    & 4 & 0.678 & 0.640 & 0.159 \\
    & 3 & 0.690$^{*}$ & 0.600 & 0.181 \\
\hline

\multirow{3}{*}{Centralized}
    & 5 & 0.746 & 0.720 & 0.245 \\
    & 4 & 0.722 & 0.688 & 0.271 \\
    & 3 & 0.724 & 0.648 & 0.318 \\
\hline

\end{tabular}

\vspace{1mm}

\footnotesize
\textit{Note:} $^{*}$Values do not exhibit statistically significant differences with Centralized according to Dunn's post-hoc test ($p > 0.05$).
\end{table}

\begin{figure*}[!htbp]
\centering
\includegraphics[width=0.9\textwidth]{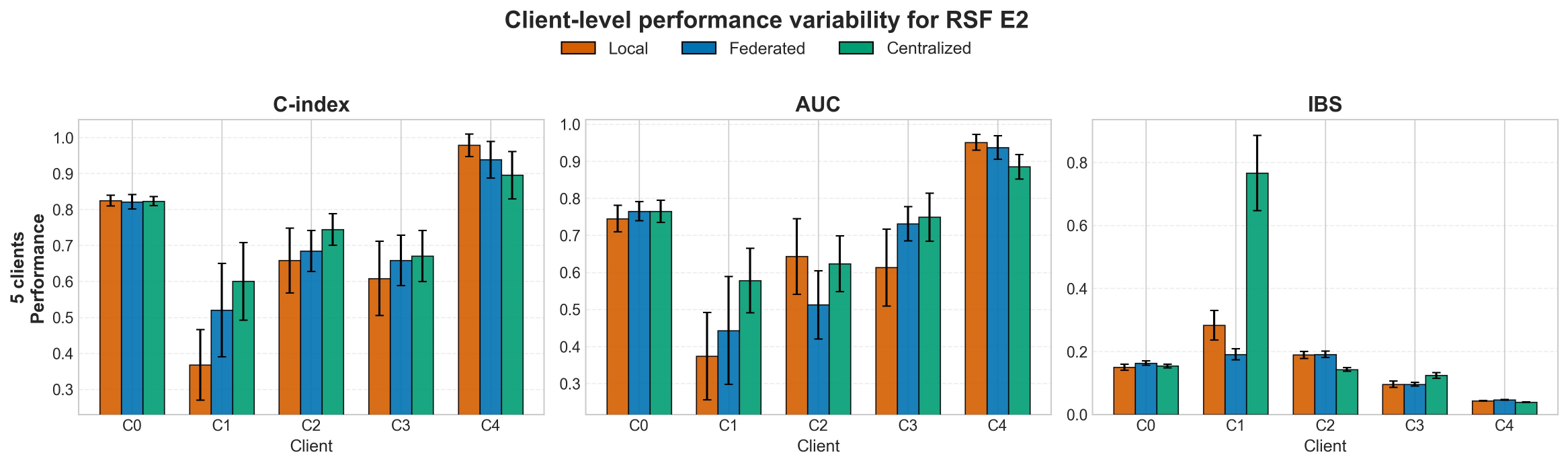}
\caption{Client-level performance variability across training paradigms for RSF under the five-client configuration (100 trees). Results are reported as mean $\pm$ standard deviation across 10 independent runs for each client: (a) C-Index, (b) AUC, and (c) IBS.}
\label{fig:rsfE2_clients}
\end{figure*}

Fig.~\ref{fig:rsfE2_clients} shows the client-level performance variability of the RSF algorithm under the five-client configuration with 100 trees. Differences across clients can be observed for all metrics and training paradigms. Client C4 obtained the highest C-Index and AUC values across all paradigms, while lower discrimination performance was observed for client C1. In terms of IBS, higher values were obtained by client C1, especially under the centralized training method. The error bars indicate variability across runs, which was more pronounced for client C1 across the discrimination metrics. Lastly, federated training achieved higher discrimination performance than local training for several clients, particularly for clients C1 and C3 in terms of C-Index and AUC.

\subsubsection{Cross-Model Comparison under Federated Learning}
\label{sec:cross_model_comparison}
The cross-model comparison is performed with a focus on FL, as this is considered to be the main scenario of interest for this work. By comparing the models in this scenario, it is possible to assess their efficiency in a real environment, as well as their robustness with respect to data heterogeneity and splitting.

\begin{figure*}[t]
\centering
\includegraphics[width=0.9\textwidth]{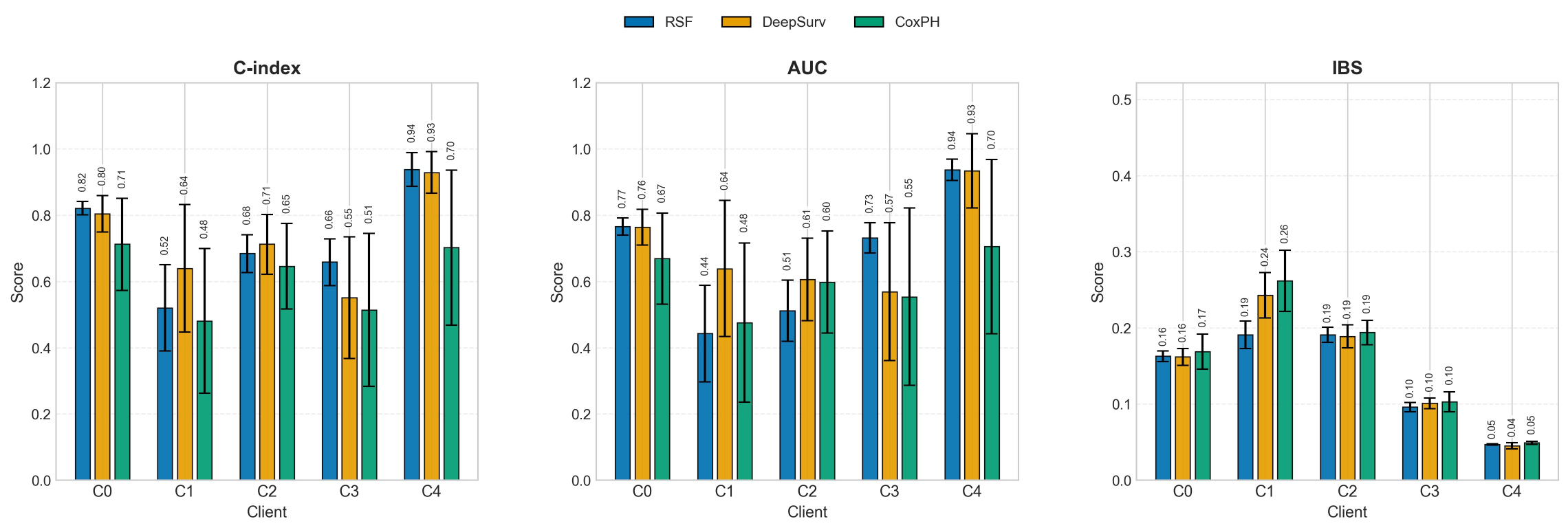}
\caption{Client-level performance comparison of RSF, DeepSurv, and CoxPH under federated learning for the five-client configuration. Results are reported as mean $\pm$ standard deviation across 10 independent runs for each client: (a) C-Index, (b) AUC, and (c) IBS.}
\label{fig:figure_modelComparison}
\end{figure*}

In order to compare the efficiency of the evaluated models, it is necessary to consider different numbers of clients (i.e., 3, 4, and 5). For a better understanding, Fig.~\ref{fig:figure_modelComparison} shows a graphical representation of the performance for 5 clients. This choice has been made as it represents the scenario with the highest number of participating clients, for whom the performance of the models tends to be higher. The performance for 3 and 4 clients has similar characteristics, as presented in Table~\ref{tab:federated_mean_comparison}.

As demonstrated in Fig.~\ref{fig:figure_modelComparison}, performance variations between the models are observed across all metrics, with the black error bars indicating standard deviations for each setup. In terms of the C-Index, RSF and DeepSurv outperform CoxPH for all clients. For clients C0 and C4, RSF has the highest C-Index, while for client C2, DeepSurv has comparable performance. For all clients, CoxPH has the lowest C-Index. Variability is higher for CoxPH for several clients, such as C1 and C3, while RSF and DeepSurv have stable performance.

The same pattern is followed by the AUC metric. RSF and DeepSurv report better results compared to CoxPH, with RSF having better results for clients C0 and C4, and DeepSurv having better results for client C1. CoxPH reports the lowest results for this metric for most clients. Regarding the standard deviation, we can see that CoxPH exhibits higher standard deviations for the clients, while RSF and DeepSurv are more consistent.

For the IBS metric, RSF reports the lowest results compared to the other methods for most of the clients, such as C1 and C3. DeepSurv reports similar results but with a slightly higher value. On the other hand, CoxPH reports higher results for this metric, especially for client C1. Regarding the standard deviation, we can see that it is smaller compared to the previous metrics, with CoxPH having a slightly higher standard deviation.

The average results reported in Table~\ref{tab:federated_mean_comparison} confirm these observations. Across all numbers of clients, RSF and DeepSurv obtain higher C-Index and AUC values than CoxPH, and the relative ordering of the models remains consistent. In terms of IBS, RSF achieves the lowest values for all configurations, followed by DeepSurv, while CoxPH consistently shows higher values.

\begin{table}[t]
\caption{Average Federated Performance Across Models and Number of Clients}
\label{tab:federated_mean_comparison}
\centering
\footnotesize
\setlength{\tabcolsep}{6pt}
\renewcommand{\arraystretch}{1.15}

\begin{tabular}{lcccc}
\hline
\textbf{Model} & \textbf{Clients} & \textbf{C-Index} ($\uparrow$) & \textbf{AUC} ($\uparrow$) & \textbf{IBS} ($\downarrow$) \\
\hline

\multirow{3}{*}{CoxPH}
    & 5 & 0.611 & 0.600 & 0.155 \\
    & 4 & 0.594 & 0.581 & 0.183 \\
    & 3 & 0.609 & 0.573 & 0.213 \\
\hline

\multirow{3}{*}{DeepSurv}
    & 5 & 0.727 & 0.702 & 0.148 \\
    & 4 & 0.671 & 0.636 & 0.174 \\
    & 3 & 0.724 & 0.667 & 0.198 \\
\hline

\multirow{3}{*}{RSF}
    & 5 & 0.724 & 0.678 & 0.138 \\
    & 4 & 0.678 & 0.640 & 0.159 \\
    & 3 & 0.690 & 0.600 & 0.181 \\
\hline

\end{tabular}

\vspace{1mm}

\footnotesize
\textit{Note:} Mean performance across clients across runs and client configurations.
\end{table}

\subsection{Performance of Federated Optimization Strategies}

The impact of different federated aggregation and optimization strategies on model performance was evaluated for the CoxPH and DeepSurv models using FedAvg, FedAdam, and FedProx. In contrast, the RSF model follows a different federated optimization approach due to its tree-based nature, relying on aggregation instead of gradient-based optimization. Therefore, the above-mentioned optimization strategies were not applicable to RSF.

Tables~\ref{tab:federated_strategies_mean_comparison_coxph}  and~\ref{tab:federated_strategies_mean_comparison_deepsurv} present the average performance obtained using different federated aggregation and optimization strategies for the CoxPH and DeepSurv models, respectively. From the results presented below, it can be observed that FedAvg and FedProx achieved comparable results across all client configurations, whereas FedAdam consistently yielded lower performance. 

Regarding the CoxPH model, FedAvg obtained the highest C-Index and AUC values for the five-client configuration, reaching 0.611 and 0.600, respectively. FedProx showed a very similar behavior, with only marginal differences across all metrics and client configurations. On the other hand, FedAdam consistently produced lower discrimination performance and slightly higher IBS values, especially for the three-client configuration, where the IBS increased to 0.217 compared to 0.213 for both FedAvg and FedProx. Taken together, the results suggest that FedAvg and FedProx provide more stable optimization behavior for CoxPH under heterogeneous client distributions in this experimental setting.

A similar trend was observed for DeepSurv, where FedAvg and FedProx achieved comparable and consistently strong performance across all client configurations. FedAvg obtained the highest C-Index and AUC values for the five-client configuration (0.727 and 0.702, respectively), whereas FedProx performed best in the three-client configuration, reaching a C-Index of 0.748 and an AUC of 0.676 and thereby outperforming both FedAvg and FedAdam. Conversely, FedAdam again produced the lowest discrimination metrics and the highest IBS values, particularly for the three-client setting. These findings indicate that proximal regularization (FedProx) can be beneficial for deep learning-based survival models under stronger client heterogeneity, while FedAvg remains a competitive and stable baseline; FedAdam, in turn, appears less robust in this experimental setting.

\begin{table}[t]
\caption{Average CoxPH Performance Across Federated Learning Strategies and Client Configurations}
\label{tab:federated_strategies_mean_comparison_coxph}
\centering
\footnotesize
\setlength{\tabcolsep}{6pt}
\renewcommand{\arraystretch}{1.15}

\begin{tabular}{lcccc}
\hline
\textbf{Strategy} & \textbf{Clients} & \textbf{C-Index} ($\uparrow$) & \textbf{AUC} ($\uparrow$) & \textbf{IBS} ($\downarrow$) \\
\hline

\multirow{3}{*}{FedAvg}
    & 5 & 0.611 & 0.600 & 0.155 \\
    & 4 & 0.594 & 0.581 & 0.183 \\
    & 3 & 0.609 & 0.573 & 0.213 \\
\hline

\multirow{3}{*}{FedAdam}
    & 5 & 0.573 & 0.553 & 0.160 \\
    & 4 & 0.561 & 0.549 & 0.188 \\
    & 3 & 0.568 & 0.543 & 0.217 \\
\hline

\multirow{3}{*}{FedProx}
    & 5 & 0.607 & 0.592 & 0.156 \\
    & 4 & 0.600 & 0.583 & 0.181 \\
    & 3 & 0.605 & 0.574 & 0.213 \\
\hline

\end{tabular}

\vspace{1mm}

\footnotesize
\textit{Note:} Mean performance across clients across runs and client configurations.
\end{table}

\begin{table}[t]
\caption{Average DeepSurv Performance Across Federated Learning Strategies and Client Configurations}
\label{tab:federated_strategies_mean_comparison_deepsurv}
\centering
\footnotesize
\setlength{\tabcolsep}{6pt}
\renewcommand{\arraystretch}{1.15}

\begin{tabular}{lcccc}
\hline
\textbf{Strategy} & \textbf{Clients} & \textbf{C-Index} ($\uparrow$) & \textbf{AUC} ($\uparrow$) & \textbf{IBS} ($\downarrow$) \\
\hline

\multirow{3}{*}{FedAvg}
    & 5 & 0.727 & 0.702 & 0.148 \\
    & 4 & 0.671 & 0.636 & 0.174 \\
    & 3 & 0.724 & 0.667 & 0.198 \\
\hline

\multirow{3}{*}{FedAdam}
    & 5 & 0.697 & 0.669 & 0.148 \\
    & 4 & 0.650 & 0.621 & 0.174 \\
    & 3 & 0.691 & 0.651 & 0.200 \\
\hline

\multirow{3}{*}{FedProx}
    & 5 & 0.719 & 0.697 & 0.148 \\
    & 4 & 0.670 & 0.637 & 0.173 \\
    & 3 & 0.748 & 0.676 & 0.197 \\
\hline

\end{tabular}

\vspace{1mm}

\footnotesize
\textit{Note:} Mean performance across clients across runs and client configurations.
\end{table}

\section{Discussion}\label{sec:discussion}

This section discusses the main findings extracted from the experimental evaluation of the proposed federated survival analysis framework. First, a comparative analysis of the different training paradigms and survival models is presented. Then, the impact of federated optimization strategies on model performance and robustness is examined. Following this, the influence of data heterogeneity and client-level variability is analyzed. Lastly, practical guidelines for federated survival modeling are provided based on the observed experimental behavior.

\subsection{Comparative Analysis of Training Paradigms and Survival Models}
\label{sec:fam_comp_disc}

A comparison between local, federated, and centralized training strategies shows that there are significant differences among the training techniques in terms of \textit{how} the survival models learn from distributed data and generalize over clients. Centralized training generally provides strong discrimination performance, as the model is trained on the entire dataset, allowing it to capture global correlations and reduce statistical variability. This behavior is consistently observed for CoxPH and RSF, where centralized training achieves the highest or near-highest performance in terms of C-Index and AUC (see Tables~\ref{tab:coxph_mean} and~\ref{tab:rsf_mean}). However, this trend is not uniform across all families. For DeepSurv, FL achieves comparable or even superior discrimination performance (see Table~\ref{tab:deepsurv_mean}). This may be attributed to the learning characteristics of Deep Neural Networks (DNN). In contrast to classical models, the DeepSurv model can benefit from the dynamics of federated training, where the aggregation of heterogeneous client updates introduces additional variance in the optimization process \cite{Li2020FedAvgNonIID}. This stochasticity can have a regularizing effect and may help improve generalization, therefore allowing federated models to outperform centralized training in certain cases.

On the other hand, FL provides a trade-off between performance and privacy by enabling collaborative training without sharing raw data. Although it cannot directly use the entire dataset, it leverages information obtained from different clients via parameter aggregation. 
As shown in Tables~\ref{tab:coxph_mean},~\ref{tab:deepsurv_mean}, and~\ref{tab:rsf_mean}, federated models generally outperform local models in terms of C-Index and AUC, benefiting from exposure to a broader range of data distributions, as reflected in Table~\ref{tab:fed_per_client}. Nevertheless, the ability to perform well is dependent on the similarity between client distributions, since if the datasets are diverse, the federated model might fail to adequately account for the particularities of the individual clients. Thus, performance will vary across clients, as depicted in Figures~\ref{fig:coxph_clients}, \ref{fig:deepsurv_clients}, and \ref{fig:rsfE2_clients}.

Local learning generally yields the lowest performance among the three training approaches. This result was expected, since each model is trained using a limited amount of data and does not benefit from shared information across clients. As shown in Table \ref{tab:local_per_client}, local models are more likely to overfit to the specific characteristics of the client and may fail to generalize effectively. This problem is especially evident when clients have scarce data and skewed distributions.

Furthermore, the comparative analysis of the different models demonstrates notable differences in their ability to handle heterogeneous datasets in federated environments. In particular, RSF and DeepSurv achieved better discrimination performance than CoxPH across most evaluated configurations. This behavior may be explained by the higher representational capacity of these models, which enables them to capture complex nonlinear relationships and interactions between covariates and survival outcomes. In contrast, CoxPH relies on proportional hazards assumptions and linear risk modeling, which limits its flexibility when client data distributions vary substantially.

Regarding calibration, RSF generally provided the most reliable survival probability estimates, followed by DeepSurv, whereas CoxPH tended to exhibit greater variability and instability. This behavior highlights the robustness of tree-based ensemble methods under heterogeneous data conditions. Although DeepSurv demonstrated strong predictive capability, its calibration performance was more sensitive to optimization dynamics and data imbalance. Similarly, the dependence of CoxPH on globally consistent risk sets may increase its sensitivity to distributional imbalances across clients. In terms of robustness, RSF showed the most stable performance across clients and training paradigms, likely due to the variance-reduction properties of ensemble learning methods. DeepSurv also achieved strong results, although with slightly higher variability caused by its dependence on optimization and initialization. In contrast, CoxPH appeared to be more sensitive to distributional shifts, resulting in greater variability across client configurations.

Overall, these findings indicate that model selection plays a critical role in federated survival analysis. RSF provided the best balance between discrimination, calibration, and robustness across distributed settings, whereas DeepSurv offered strong predictive capability with moderate sensitivity to optimization dynamics. CoxPH remained advantageous primarily in more homogeneous and interpretability-oriented scenarios. These results highlight a clear trade-off between data availability and the efficiency of the models being trained. Centralized training remains the most effective solution when data centralization is possible, owing to its superior ability to learn global patterns. FL provides a strong alternative under privacy constraints, improving over local training while maintaining data decentralization, although its effectiveness is influenced by client heterogeneity. Local training, while privacy-preserving, is limited by data scarcity and lack of generalization.

\subsection{Impact of Federated Optimization Strategies}

The impact of the federated optimization strategies was assessed for the two gradient-based models, CoxPH and DeepSurv, since RSF aggregates decision trees rather than gradient updates and is therefore incompatible with these strategies. As summarized in Table~\ref{tab:federated_strategy_summary} and detailed in Tables~\ref{tab:federated_strategies_mean_comparison_coxph} and~\ref{tab:federated_strategies_mean_comparison_deepsurv}, FedAvg and FedProx provided the most competitive and stable performance across all client configurations, whereas FedAdam consistently yielded inferior discrimination performance.

For the CoxPH model, FedAvg and FedProx showed very similar behavior across all evaluated configurations, indicating that both strategies are appropriate choices for federated implementations of traditional statistical survival models. On the other hand, the lower performance observed with FedAdam suggests that adaptive \textit{server-side} optimization did not provide clear advantages under the evaluated experimental conditions, which is consistent with the limited benefit that adaptive methods typically offer for the convex, low-dimensional optimization underlying the Cox partial likelihood.

For DeepSurv, FedProx and FedAvg were closely matched and clearly ahead of FedAdam. FedAvg obtained the best discrimination performance in the five-client configuration, whereas FedProx achieved the highest overall scores in the three-client configuration (a C-Index of 0.748 and an AUC of 0.676). The proximal term used by FedProx is designed to limit client drift when local updates diverge under heterogeneous data, consistent with its characterization in Table~\ref{tab:federated_strategy_summary}; in this study, however, its advantage over a well-tuned FedAvg remained marginal.

Altogether, these findings suggest that simpler aggregation strategies such as FedAvg can already achieve competitive performance for federated survival analysis, while FedProx can provide additional robustness for more complex neural-network-based models when client distributions diverge. In contrast, FedAdam did not demonstrate clear advantages in the evaluated experimental scenarios and would likely require careful hyperparameter tuning to become competitive.

\begin{table}[t]
\caption{Summary of the Evaluated Federated Optimization Strategies}
\label{tab:federated_strategy_summary}
\centering
\footnotesize
\setlength{\tabcolsep}{4pt}
\renewcommand{\arraystretch}{1.3}

\begin{tabular}{@{}l p{2.75cm} p{3.3cm}@{}}
\hline
\textbf{Strategy} & \textbf{Aggregation Mechanism} & \textbf{Behavior in This Study} \\
\hline
FedAvg~\cite{mcmahan2017communication}  & Weighted averaging of local model updates & Stable and competitive baseline across all configurations \\
FedProx~\cite{li2020federated} & Adds a proximal term that limits client drift & Comparable to FedAvg, with occasional gains under heterogeneity \\
FedAdam~\cite{reddi2020adaptive} & Adaptive server-side optimization & Consistently lower discrimination performance \\
\hline
\end{tabular}
\end{table}

\subsection{Data Heterogeneity}
The effects of heterogeneity in the data are studied by varying the number of participating clients, thereby effectively limiting the set of distributions considered for the training procedure. In this case, each client keeps its original distribution, while the reduction in the number of clients implies the exclusion of certain data distributions from consideration rather than redistributing data across clients. An interesting result is that the performance of the model does not decrease monotonically with a decrease in the number of clients. This can be observed across all models, as shown in Tables~\ref{tab:coxph_mean},~\ref{tab:deepsurv_mean}, and~\ref{tab:rsf_mean}, where no consistent improvement or degradation is observed when moving from 5 to 3 clients. This indicates that the performance of the model does not depend on the number of clients themselves but rather on the type of data distribution used for training the model. With an increase in the number of clients, a greater variety of data in terms of feature distribution, survival pattern, and censoring is taken into account.

This becomes especially apparent from the way calibration behaves. As fewer clients participate in the training, calibration performance degrades for all training modes, as reflected by the generally increasing IBS values in Tables~\ref{tab:coxph_mean}, \ref{tab:deepsurv_mean}, and \ref{tab:rsf_mean}. This might suggest that the estimation of survival probabilities is sensitive to the diversity of observed survival times and event patterns. When certain clients are excluded, the model may fail to capture important regions of the time-to-event distribution, which can lead to less reliable probability estimates. On the other hand, the discrimination task performs quite consistently regardless of whether clients differ in their structure. As shown in Tables~\ref{tab:coxph_mean}, \ref{tab:deepsurv_mean}, and \ref{tab:rsf_mean}, C-Index and AUC values exhibit only minor fluctuations when the number of clients changes. The reason for this may be that ranking clients based on their risk is less sensitive to the presence or absence of specific data distributions. Ranking depends on the relative ordering of individual risks, which can still be recovered even when the diversity of the data is reduced. Calibration, on the other hand, depends more on capturing the full underlying distribution.

In addition to this, the variations seen amongst the clients also point to the differences present in the distribution of data for the individual clients. As shown in Table~\ref{tab:fed_per_client}, some clients consistently achieve better performance, whereas others display poorer performance and greater variability. A notable exception is observed for client C1, where centralized training shows significantly higher IBS values. This behavior is evident in Table~\ref{tab:centralized_per_client}. This could be because longer survival times exist in client C1, as it includes some of the highest event times that have been observed. Specifically, there are very late event occurrences in client C1 that have occurred beyond 4000 days, but do not exist in the test data. Because centralized learning aggregates information from all clients, the model learns to assign a nonzero risk score at much later time points than the other clients, leading to poorer calibration in this case. A similar behavior is also observed for the CoxPH and DeepSurv models (see Sections~\ref{sec:coxph_results} and ~\ref{sec:deepsurv_results}), indicating that this effect is not specific to the RSF model, but rather related to the underlying data distribution. Therefore, it can be suggested that different characteristics such as sample size, censoring ratio, and feature distribution vary among the clients. In such scenarios, excluding specific clients may significantly affect the global model, particularly if they contain unique or extreme patterns.

Moreover, the observed differences between clients suggest that the specific properties of the data, such as censoring levels, event scarcity, and institutional variance, play a critical role when evaluating model performance. Clients that have few events observed or exhibit high censoring ratios contribute less information to the learning process, which can result in poor discrimination and calibration of the model. In a similar way, institutional imbalance due to unequal distribution of data among different clients may result in a biased global model. In summary, the influence of data heterogeneity appears to depend more on the diversity of the participating clients and how well they represent different types of data distributions, and not necessarily their number.

\subsection{Practical Guidelines}
Based on the results obtained through the different experiments, certain practical recommendations can be drawn regarding how to choose proper models and how to conduct training when conducting federated survival analysis. These recommendations are summarized in Table~\ref{tab:practical_guidelines} and discussed below, with each guideline grounded in the behavior observed under the different data and training conditions.

\begin{table*}[t]
\caption{Practical Guidelines for Federated Survival Modeling}
\label{tab:practical_guidelines}
\centering
\footnotesize
\renewcommand{\arraystretch}{1.2}
\setlength{\tabcolsep}{8pt}

\begin{tabular}{llll}
\hline
\textbf{Scenario} & \textbf{Model} & \textbf{Training} & \textbf{Rationale} \\
\hline

Heterogeneous clients 
& RSF 
& Federated
& Robust to distribution shifts and stable across clients \\

Homogeneous data 
& CoxPH 
& Centralized or Federated 
& Linear model; assumptions are likely valid \\

Scarce data 
& RSF or DeepSurv 
& Federated
& Aggregation improves generalization across clients \\

Calibration-critical applications 
& RSF 
& Federated
& More reliable survival probability estimates \\

Interpretability-focused studies 
& CoxPH 
& Centralized or Federated 
& Interpretable coefficients and hazard ratios \\

Strict privacy constraints 
& RSF or DeepSurv 
& Federated
& Collaborative learning without raw data sharing \\

Limited computational resources 
& CoxPH 
& Local or Federated 
& Computationally efficient with fast training \\
\hline

\end{tabular}
\end{table*}

\paragraph{Heterogeneous clients} When client distributions differ substantially, RSF with federated training is recommended. This is supported by the results in Section~\ref{sec:results_rsf} and the cross-model comparison (Section~\ref{sec:cross_model_comparison}), where RSF consistently shows stable performance across clients and lower variability compared to CoxPH and DeepSurv. In particular, RSF maintains competitive discrimination while showing robustness to distribution shifts, as demonstrated by the relatively small performance gaps between federated and centralized training (Table~\ref{tab:rsf_mean}) and its stability across clients in Figure~\ref{fig:rsfE2_clients}.

\paragraph{Homogeneous data} For homogeneous datasets, CoxPH is a suitable choice, since its underlying assumptions are more likely to hold. As discussed in Section~\ref{sec:cross_model_comparison}, CoxPH performs suitably when data distributions are consistent, but its performance degrades under heterogeneity due to its linear structure and reliance on proportional hazards. This is also reflected in the higher variability observed across clients in the CoxPH blocks of Tables~\ref{tab:fed_per_client}, \ref{tab:local_per_client}, and \ref{tab:centralized_per_client}.

\paragraph{Scarce or imbalanced data} When data are scarce or imbalanced across clients, FL with RSF or DeepSurv is preferable. The results of the experiments presented in Sections~\ref{sec:coxph_results} and~\ref{sec:deepsurv_results} show that federated training improves performance over local models, particularly for weaker clients such as C1, where noticeable gains in C-Index and AUC are observed (Figures~\ref{fig:coxph_clients} and~\ref{fig:deepsurv_clients}). This indicates that aggregation helps mitigate data scarcity by leveraging information from other clients.

\paragraph{Calibration-critical applications} When calibration is critical, RSF with federated training should be chosen. According to Table~\ref{tab:federated_mean_comparison} from Section~\ref{sec:cross_model_comparison}, RSF has the lowest IBS among all options considered, thus demonstrating its superior accuracy in survival probability estimation compared with the other models, such as DeepSurv and CoxPH.

\paragraph{Interpretability-focused studies} When interpretability matters, CoxPH remains the preferred model. As discussed in Section~\ref{sec:fam_comp_disc}, it provides interpretable coefficients and hazard ratios, making it well suited for clinical settings despite its lower predictive performance.

\paragraph{Strict privacy constraints} When strict privacy requirements must be met, FL with RSF or DeepSurv is recommended. All the results from different experiments show that the federated model is better than the local one and that it can approximate centralized performance (Sections~\ref{sec:deepsurv_results} and~\ref{sec:results_rsf}), therefore making them a practical solution when data sharing is not possible.

\paragraph{Limited computational resources} Finally, in environments with limited computational resources, CoxPH is a practical option due to its lower computational complexity. This is due to the fact that as a linear model, it requires less training time and fewer resources compared to RSF and DeepSurv, while still providing acceptable performance in simpler scenarios.

Nonetheless, it is important to note that this discussion is specific to the dataset used in this study, and the comparative effectiveness of the models may differ under other data characteristics, such as variable distributions, censoring effects, and client heterogeneity, as well as application constraints like interpretability, privacy, and computational cost.

\section{Conclusion}
\label{sec:conclusion}

This paper presented a systematic, multi-model evaluation of federated survival analysis on the Fed-TCGA-BRCA dataset, a cross-institutional breast cancer cohort with naturally heterogeneous clients. Federated training consistently outperformed local training and approached centralized performance without sharing raw data, occasionally exceeding it for DeepSurv, where aggregation of heterogeneous updates appears to act as a regularizer. Among the models, RSF offered the best overall balance of discrimination, calibration, and robustness across clients, whereas CoxPH remained competitive mainly in homogeneous and interpretability-oriented settings. Performance was governed by the diversity of client distributions rather than their number, and FedAvg and FedProx proved more stable than FedAdam in this setting. Building on these observations, the study's main practical contribution is a set of decision-oriented guidelines that map data, privacy, interpretability, and resource constraints to suitable model and paradigm choices for federated survival modeling in privacy-constrained healthcare environments.

Several directions remain open. Larger federations with more data could improve stability and calibration, which was most sensitive to client exclusion, while extreme or unbalanced distributions motivate data-balancing strategies and more robust estimators. Personalized or hybrid federated schemes may better accommodate strong heterogeneity, and a closer analysis of communication efficiency would clarify scalability to larger deployments. Finally, integrating formal privacy-preserving mechanisms such as differential privacy, and improving the interpretability of RSF and DeepSurv, would further strengthen the clinical adoption of federated survival models.

\appendices

\section{Client-Level Performance Results}
\label{app:client_tables}
This appendix reports the detailed per-client performance results that complement the aggregated values discussed in the main text. All values are reported as mean $\pm$ standard deviation over 10 independent runs. Table~\ref{tab:local_per_client} reports the per-client performance under local training, identical regardless of the number of participating clients since each client is trained independently; Table~\ref{tab:fed_per_client} reports the per-client performance under federated training; and Table~\ref{tab:centralized_per_client} reports the per-client performance under centralized training.

\begin{table}[t]
\caption{Local Per-Client Performance}
\label{tab:local_per_client}
\centering
\footnotesize
\setlength{\tabcolsep}{6pt}
\renewcommand{\arraystretch}{1.15}

\begin{tabular}{lccc}
\hline
\textbf{Client} &
\textbf{C-Index} ($\uparrow$) &
\textbf{AUC} ($\uparrow$) &
\textbf{IBS} ($\downarrow$)
\\
\hline
\multicolumn{4}{c}{\textbf{CoxPH}} \\
\hline
\begin{tabular}[c]{@{}l@{}}
C0 \\
C1 \\
C2 \\
C3 \\
C4
\end{tabular}
&
\begin{tabular}[c]{@{}l@{}}
0.784 $\pm$ 0.045 \\
0.350 $\pm$ 0.156 \\
0.577 $\pm$ 0.131 \\
0.552 $\pm$ 0.221 \\
0.725 $\pm$ 0.254
\end{tabular}
&
\begin{tabular}[c]{@{}l@{}}
0.718 $\pm$ 0.054 \\
0.305 $\pm$ 0.158 \\
0.640 $\pm$ 0.134 \\
0.594 $\pm$ 0.248 \\
0.771 $\pm$ 0.275
\end{tabular}
&
\begin{tabular}[c]{@{}l@{}}
0.169 $\pm$ 0.018 \\
0.314 $\pm$ 0.052 \\
0.198 $\pm$ 0.018 \\
0.106 $\pm$ 0.019 \\
0.047 $\pm$ 0.004
\end{tabular}
\\
\hline
\multicolumn{4}{c}{\textbf{DeepSurv}} \\
\hline

\begin{tabular}[c]{@{}l@{}}
C0 \\
C1 \\
C2 \\
C3 \\
C4
\end{tabular}
&
\begin{tabular}[c]{@{}l@{}}
0.740 $\pm$ 0.130 \\
0.521 $\pm$ 0.166 \\
0.543 $\pm$ 0.118 \\
0.509 $\pm$ 0.119 \\
0.775 $\pm$ 0.161
\end{tabular}
&
\begin{tabular}[c]{@{}l@{}}
0.689 $\pm$ 0.102 \\
0.515 $\pm$ 0.163 \\
0.582 $\pm$ 0.120 \\
0.510 $\pm$ 0.124 \\
0.807 $\pm$ 0.182
\end{tabular}
&
\begin{tabular}[c]{@{}l@{}}
0.179 $\pm$ 0.015 \\
0.270 $\pm$ 0.047 \\
0.195 $\pm$ 0.011 \\
0.103 $\pm$ 0.007 \\
0.046 $\pm$ 0.005
\end{tabular}
\\
\hline
\multicolumn{4}{c}{\textbf{RSF}} \\
\hline
\begin{tabular}[c]{@{}l@{}}
C0 \\
C1 \\
C2 \\
C3 \\
C4
\end{tabular}
&
\begin{tabular}[c]{@{}l@{}}
0.824 $\pm$ 0.015 \\
0.368 $\pm$ 0.098 \\
0.658 $\pm$ 0.090 \\
0.608 $\pm$ 0.103 \\
0.978 $\pm$ 0.031
\end{tabular}
&
\begin{tabular}[c]{@{}l@{}}
0.745 $\pm$ 0.036 \\
0.374 $\pm$ 0.118 \\
0.643 $\pm$ 0.102 \\
0.613 $\pm$ 0.104 \\
0.951 $\pm$ 0.021
\end{tabular}
&
\begin{tabular}[c]{@{}l@{}}
0.150 $\pm$ 0.010 \\
0.283 $\pm$ 0.047 \\
0.189 $\pm$ 0.011 \\
0.096 $\pm$ 0.010 \\
0.043 $\pm$ 0.001
\end{tabular}
\\
\hline

\end{tabular}

\vspace{1mm}

\footnotesize
\textit{Note:} Reported values correspond to mean $\pm$ std. deviation over 10 runs.
\end{table}

\begin{table*}[t]
\caption{Federated Per-Client Performance Across Different Numbers of Clients}
\label{tab:fed_per_client}
\centering
\footnotesize
\setlength{\tabcolsep}{3pt}
\renewcommand{\arraystretch}{1.15}
\begin{tabular}{lccccccccc}
\hline
\textbf{Client} &
\multicolumn{3}{c}{\textbf{CoxPH (FedAvg)}} &
\multicolumn{3}{c}{\textbf{DeepSurv (FedAvg)}} &
\multicolumn{3}{c}{\textbf{RSF}} \\
\cmidrule(lr){2-4}\cmidrule(lr){5-7}\cmidrule(lr){8-10}
& \textbf{C-Index} ($\uparrow$) & \textbf{AUC} ($\uparrow$) & \textbf{IBS} ($\downarrow$) &
  \textbf{C-Index} ($\uparrow$) & \textbf{AUC} ($\uparrow$) & \textbf{IBS} ($\downarrow$) &
  \textbf{C-Index} ($\uparrow$) & \textbf{AUC} ($\uparrow$) & \textbf{IBS} ($\downarrow$) \\
\hline
\multicolumn{10}{c}{\textit{5 Clients}} \\
\hline
C0 & 0.712 $\pm$ 0.139 & 0.669 $\pm$ 0.137 & 0.169 $\pm$ 0.023 & 0.804 $\pm$ 0.055 & 0.763 $\pm$ 0.054 & 0.162 $\pm$ 0.011 & 0.821 $\pm$ 0.020 & 0.765 $\pm$ 0.026 & 0.163 $\pm$ 0.007 \\
C1 & 0.481 $\pm$ 0.218 & 0.476 $\pm$ 0.240 & 0.262 $\pm$ 0.040 & 0.640 $\pm$ 0.192 & 0.639 $\pm$ 0.205 & 0.243 $\pm$ 0.030 & 0.520 $\pm$ 0.130 & 0.443 $\pm$ 0.146 & 0.191 $\pm$ 0.018 \\
C2 & 0.646 $\pm$ 0.129 & 0.598 $\pm$ 0.154 & 0.194 $\pm$ 0.016 & 0.712 $\pm$ 0.090 & 0.606 $\pm$ 0.124 & 0.189 $\pm$ 0.015 & 0.684 $\pm$ 0.057 & 0.512 $\pm$ 0.092 & 0.191 $\pm$ 0.010 \\
C3 & 0.514 $\pm$ 0.231 & 0.554 $\pm$ 0.268 & 0.103 $\pm$ 0.013 & 0.551 $\pm$ 0.183 & 0.569 $\pm$ 0.208 & 0.101 $\pm$ 0.007 & 0.658 $\pm$ 0.070 & 0.731 $\pm$ 0.046 & 0.096 $\pm$ 0.006 \\
C4 & 0.702 $\pm$ 0.234 & 0.705 $\pm$ 0.263 & 0.049 $\pm$ 0.002 & 0.929 $\pm$ 0.063 & 0.934 $\pm$ 0.112 & 0.045 $\pm$ 0.004 & 0.938 $\pm$ 0.051 & 0.937 $\pm$ 0.032 & 0.047 $\pm$ 0.001 \\
\hline
\multicolumn{10}{c}{\textit{4 Clients}} \\
\hline
C0 & 0.703 $\pm$ 0.125 & 0.659 $\pm$ 0.123 & 0.171 $\pm$ 0.022 & 0.803 $\pm$ 0.025 & 0.753 $\pm$ 0.021 & 0.164 $\pm$ 0.013 & 0.804 $\pm$ 0.020 & 0.756 $\pm$ 0.026 & 0.161 $\pm$ 0.005 \\
C1 & 0.481 $\pm$ 0.241 & 0.471 $\pm$ 0.267 & 0.266 $\pm$ 0.042 & 0.615 $\pm$ 0.158 & 0.604 $\pm$ 0.170 & 0.246 $\pm$ 0.031 & 0.511 $\pm$ 0.146 & 0.468 $\pm$ 0.162 & 0.194 $\pm$ 0.016 \\
C2 & 0.666 $\pm$ 0.123 & 0.635 $\pm$ 0.142 & 0.192 $\pm$ 0.016 & 0.728 $\pm$ 0.085 & 0.627 $\pm$ 0.131 & 0.185 $\pm$ 0.013 & 0.730 $\pm$ 0.028 & 0.583 $\pm$ 0.078 & 0.184 $\pm$ 0.010 \\
C3 & 0.527 $\pm$ 0.250 & 0.560 $\pm$ 0.280 & 0.102 $\pm$ 0.013 & 0.539 $\pm$ 0.147 & 0.561 $\pm$ 0.164 & 0.098 $\pm$ 0.008 & 0.668 $\pm$ 0.043 & 0.753 $\pm$ 0.043 & 0.096 $\pm$ 0.007 \\
\hline
\multicolumn{10}{c}{\textit{3 Clients}} \\
\hline
C0 & 0.735 $\pm$ 0.107 & 0.692 $\pm$ 0.101 & 0.168 $\pm$ 0.020 & 0.808 $\pm$ 0.025 & 0.751 $\pm$ 0.024 & 0.165 $\pm$ 0.013 & 0.807 $\pm$ 0.011 & 0.749 $\pm$ 0.034 & 0.159 $\pm$ 0.007 \\
C1 & 0.445 $\pm$ 0.218 & 0.408 $\pm$ 0.234 & 0.278 $\pm$ 0.041 & 0.610 $\pm$ 0.166 & 0.592 $\pm$ 0.194 & 0.249 $\pm$ 0.040 & 0.554 $\pm$ 0.127 & 0.475 $\pm$ 0.133 & 0.197 $\pm$ 0.021 \\
C2 & 0.647 $\pm$ 0.126 & 0.620 $\pm$ 0.132 & 0.192 $\pm$ 0.015 & 0.754 $\pm$ 0.063 & 0.657 $\pm$ 0.117 & 0.180 $\pm$ 0.015 & 0.708 $\pm$ 0.046 & 0.575 $\pm$ 0.085 & 0.186 $\pm$ 0.010 \\
\hline
\end{tabular}

\vspace{1mm}
\footnotesize
\textit{Note:} Reported values correspond to mean $\pm$ standard deviation over 10 runs.
\end{table*}

\begin{table*}[t]
\caption{Centralized Per-Client Performance Across Different Numbers of Clients}
\label{tab:centralized_per_client}
\centering
\footnotesize
\setlength{\tabcolsep}{3pt}
\renewcommand{\arraystretch}{1.15}
\begin{tabular}{lccccccccc}
\hline
\textbf{Client} &
\multicolumn{3}{c}{\textbf{CoxPH}} &
\multicolumn{3}{c}{\textbf{DeepSurv}} &
\multicolumn{3}{c}{\textbf{RSF}} \\
\cmidrule(lr){2-4}\cmidrule(lr){5-7}\cmidrule(lr){8-10}
& \textbf{C-Index} ($\uparrow$) & \textbf{AUC} ($\uparrow$) & \textbf{IBS} ($\downarrow$) &
  \textbf{C-Index} ($\uparrow$) & \textbf{AUC} ($\uparrow$) & \textbf{IBS} ($\downarrow$) &
  \textbf{C-Index} ($\uparrow$) & \textbf{AUC} ($\uparrow$) & \textbf{IBS} ($\downarrow$) \\
\hline
\multicolumn{10}{c}{\textit{5 Clients}} \\
\hline
C0 & 0.873 $\pm$ 0.016 & 0.832 $\pm$ 0.031 & 0.146 $\pm$ 0.008 & 0.773 $\pm$ 0.145 & 0.737 $\pm$ 0.129 & 0.167 $\pm$ 0.019 & 0.823 $\pm$ 0.013 & 0.765 $\pm$ 0.030 & 0.154 $\pm$ 0.006 \\
C1 & 0.590 $\pm$ 0.131 & 0.600 $\pm$ 0.124 & 0.965 $\pm$ 0.184 & 0.723 $\pm$ 0.149 & 0.724 $\pm$ 0.162 & 0.841 $\pm$ 0.225 & 0.600 $\pm$ 0.108 & 0.578 $\pm$ 0.087 & 0.766 $\pm$ 0.119 \\
C2 & 0.666 $\pm$ 0.093 & 0.583 $\pm$ 0.129 & 0.160 $\pm$ 0.016 & 0.616 $\pm$ 0.147 & 0.541 $\pm$ 0.202 & 0.158 $\pm$ 0.016 & 0.744 $\pm$ 0.044 & 0.623 $\pm$ 0.075 & 0.143 $\pm$ 0.006 \\
C3 & 0.581 $\pm$ 0.181 & 0.521 $\pm$ 0.211 & 0.143 $\pm$ 0.017 & 0.570 $\pm$ 0.155 & 0.550 $\pm$ 0.195 & 0.141 $\pm$ 0.010 & 0.670 $\pm$ 0.071 & 0.749 $\pm$ 0.065 & 0.124 $\pm$ 0.009 \\
C4 & 0.948 $\pm$ 0.030 & 0.951 $\pm$ 0.034 & 0.038 $\pm$ 0.005 & 0.851 $\pm$ 0.188 & 0.869 $\pm$ 0.198 & 0.040 $\pm$ 0.004 & 0.895 $\pm$ 0.066 & 0.885 $\pm$ 0.033 & 0.039 $\pm$ 0.001 \\
\hline
\multicolumn{10}{c}{\textit{4 Clients}} \\
\hline
C0 & 0.855 $\pm$ 0.028 & 0.803 $\pm$ 0.040 & 0.135 $\pm$ 0.011 & 0.756 $\pm$ 0.136 & 0.712 $\pm$ 0.116 & 0.152 $\pm$ 0.013 & 0.824 $\pm$ 0.015 & 0.768 $\pm$ 0.021 & 0.137 $\pm$ 0.005 \\
C1 & 0.631 $\pm$ 0.105 & 0.616 $\pm$ 0.105 & 0.769 $\pm$ 0.125 & 0.686 $\pm$ 0.179 & 0.683 $\pm$ 0.169 & 0.744 $\pm$ 0.083 & 0.559 $\pm$ 0.075 & 0.520 $\pm$ 0.071 & 0.715 $\pm$ 0.089 \\
C2 & 0.733 $\pm$ 0.055 & 0.661 $\pm$ 0.086 & 0.138 $\pm$ 0.013 & 0.694 $\pm$ 0.115 & 0.636 $\pm$ 0.121 & 0.138 $\pm$ 0.009 & 0.786 $\pm$ 0.037 & 0.675 $\pm$ 0.065 & 0.130 $\pm$ 0.007 \\
C3 & 0.589 $\pm$ 0.183 & 0.554 $\pm$ 0.208 & 0.112 $\pm$ 0.014 & 0.571 $\pm$ 0.187 & 0.571 $\pm$ 0.214 & 0.116 $\pm$ 0.012 & 0.718 $\pm$ 0.058 & 0.789 $\pm$ 0.046 & 0.101 $\pm$ 0.007 \\
\hline
\multicolumn{10}{c}{\textit{3 Clients}} \\
\hline
C0 & 0.861 $\pm$ 0.020 & 0.807 $\pm$ 0.030 & 0.141 $\pm$ 0.008 & 0.717 $\pm$ 0.154 & 0.677 $\pm$ 0.131 & 0.163 $\pm$ 0.012 & 0.832 $\pm$ 0.018 & 0.770 $\pm$ 0.030 & 0.140 $\pm$ 0.004 \\
C1 & 0.536 $\pm$ 0.103 & 0.549 $\pm$ 0.073 & 0.810 $\pm$ 0.153 & 0.584 $\pm$ 0.132 & 0.594 $\pm$ 0.117 & 0.730 $\pm$ 0.084 & 0.579 $\pm$ 0.120 & 0.502 $\pm$ 0.093 & 0.682 $\pm$ 0.099 \\
C2 & 0.700 $\pm$ 0.086 & 0.619 $\pm$ 0.139 & 0.152 $\pm$ 0.020 & 0.649 $\pm$ 0.106 & 0.578 $\pm$ 0.102 & 0.148 $\pm$ 0.016 & 0.762 $\pm$ 0.044 & 0.671 $\pm$ 0.063 & 0.132 $\pm$ 0.005 \\
\hline
\end{tabular}

\vspace{1mm}
\footnotesize
\textit{Note:} Reported values correspond to mean $\pm$ standard deviation over 10 runs.
\end{table*}

\section*{References}
\bibliographystyle{IEEEtran}
\bibliography{References}

@article{fernandez2025differential,
  title={Differential Privacy: Gradient Leakage Attacks in Federated Learning Environments},
  author={Fernandez-de-Retana, Miguel and Zulaika, Unai and S{\'a}nchez-Corcuera, Rub{\'e}n and Almeida, Aitor},
  journal={arXiv preprint arXiv:2510.23931},
  year={2025}
}

@article{Harrell1982,
	title = {{Evaluating the Yield of Medical Tests}},
	volume = {247},
	number = {18},
	journal = {JAMA},
	author = {Harrell Jr., F. E. and Califf, R. M. and Pryor, D. B. and Lee, K. L. and Rosati, R. A.},
	year = {1982},
	pages = {2543--2546},
	doi = {10.1001/jama.1982.03320430047030},
}

@article{Heagerty2000,
	title = {{Time-dependent {ROC} curves for censored survival data and a diagnostic marker}},
	volume = {56},
	number = {2},
	journal = {Biometrics},
	author = {Heagerty, Patrick J. and Lumley, Thomas and Pepe, Margaret S.},
	year = {2000},
	pages = {337--344},
	doi = {10.1111/j.0006-341X.2000.00337.x},
}

@article{Graf1999,
	title = {Assessment and comparison of prognostic classification schemes for survival data},
	journal = {Statistics in Medicine},
	author = {Graf, Erika and Schmoor, Claudia and Sauerbrei, Willi and Schumacher, Martin},
	year = {1999},
	volume = {18},
	number = {17-18},
	pages = {2529--2545},
	doi = {10.1002/(sici)1097-0258(19990915/30)18:17/18<2529::aid-sim274>3.0.co;2-5},
}

@article{katzman2018deepsurv,
  title={DeepSurv: personalized treatment recommender system using a Cox proportional hazards deep neural network},
  author={Katzman, Jared L and Shaham, Uri and Cloninger, Alexander and Bates, Jonathan and Jiang, Tingting and Kluger, Yuval},
  journal={BMC medical research methodology},
  volume={18},
  number={1},
  pages={24},
  year={2018},
  publisher={Springer}
}

@inproceedings{Li2020FedAvgNonIID,
	title = {{On the Convergence of FedAvg on Non-IID Data}},
	author = {Li, Xiang and Huang, Kaixuan and Yang, Wenhao and Wang, Shusen and Zhang, Zhihua},
	booktitle = {International Conference on Learning Representations (ICLR)},
	year = {2020},
}

@inproceedings{mcmahan2017communication,
  title={Communication-efficient learning of deep networks from decentralized data},
  author={McMahan, Brendan and Moore, Eider and Ramage, Daniel and Hampson, Seth and y Arcas, Blaise Aguera},
  booktitle={Artificial intelligence and statistics},
  pages={1273--1282},
  year={2017},
  organization={Pmlr}
}

@article{li2020federated,
  title={Federated optimization in heterogeneous networks},
  author={Li, Tian and Sahu, Anit Kumar and Zaheer, Manzil and Sanjabi, Maziar and Talwalkar, Ameet and Smith, Virginia},
  journal={Proceedings of Machine learning and systems},
  volume={2},
  pages={429--450},
  year={2020}
}

@article{reddi2020adaptive,
  title={Adaptive federated optimization},
  author={Reddi, Sashank and Charles, Zachary and Zaheer, Manzil and Garrett, Zachary and Rush, Keith and Kone{\v{c}}n{\`y}, Jakub and Kumar, Sanjiv and McMahan, H Brendan},
  journal={arXiv preprint arXiv:2003.00295},
  year={2020}
}

@article{topol2019high,
  title={High-performance medicine: the convergence of human and artificial intelligence},
  author={Topol, Eric J},
  journal={Nature medicine},
  volume={25},
  pages={44--56},
  year={2019},
  publisher={Nature Publishing Group US New York}
}

@article{maslej2025artificial,
  title={Artificial intelligence index report 2025},
  author={Maslej, Nestor and Fattorini, Loredana and Perrault, Raymond and Gil, Yolanda and Parli, Vanessa and Kariuki, Njenga and Capstick, Emily and Reuel, Anka and Brynjolfsson, Erik and Etchemendy, John and others},
  journal={arXiv preprint arXiv:2504.07139},
  year={2025}
}

@article{kelly2019key,
  title={Key challenges for delivering clinical impact with artificial intelligence},
  author={Kelly, Christopher J and Karthikesalingam, Alan and Suleyman, Mustafa and Corrado, Greg and King, Dominic},
  journal={BMC medicine},
  volume={17},
  number={1},
  pages={195},
  year={2019},
  publisher={Springer}
}

@article{rieke2020future,
  title={The future of digital health with federated learning},
  author={Rieke, Nicola and Hancox, Jonny and Li, Wenqi and Milletari, Fausto and Roth, Holger R and Albarqouni, Shadi and Bakas, Spyridon and Galtier, Mathieu N and Landman, Bennett A and Maier-Hein, Klaus and others},
  journal={NPJ digital medicine},
  volume={3},
  number={1},
  pages={119},
  year={2020},
  publisher={Nature Publishing Group UK London}
}

@misc{gdpr2016,
  author = {{European Commission}},
  publisher = {European Commission},
  title = {Regulation ({EU}) 2016/679 of the {European} {Parliament} and of the {Council} of 27 {April} 2016 on the protection of natural persons with regard to the processing of personal data and on the free movement of such data, and repealing {Directive} 95/46/{EC} ({General} {Data} {Protection} {Regulation})},
  url = {https://gdpr-info.eu/},
  year = 2016
}

@misc{hipaa,
  author = {U.S. Department of Health and Human Services},
  publisher = {U.S. Department of Health and Human Services},
  title = {{The Health Insurance Portability and Accountability Act of 1996 (HIPAA)}},
  year = 1996,
  url = {https://www.hhs.gov/hipaa/index.html}
}

@article{chen2021ethical,
  title={Ethical machine learning in healthcare},
  author={Chen, Irene Y and Pierson, Emma and Rose, Sherri and Joshi, Shalmali and Ferryman, Kadija and Ghassemi, Marzyeh},
  journal={Annual review of biomedical data science},
  volume={4},
  number={1},
  pages={123--144},
  year={2021},
  publisher={Annual Reviews}
}

@article{dayan2021federated,
  title={Federated learning for predicting clinical outcomes in patients with COVID-19},
  author={Dayan, Ittai and Roth, Holger R and Zhong, Aoxiao and Harouni, Ahmed and Gentili, Amilcare and Abidin, Anas Z and Liu, Andrew and Costa, Anthony Beardsworth and Wood, Bradford J and Tsai, Chien-Sung and others},
  journal={Nature medicine},
  volume={27},
  number={10},
  pages={1735--1743},
  year={2021},
  publisher={Nature Publishing Group US New York}
}

@inproceedings{li2019privacy,
  title={Privacy-preserving federated brain tumour segmentation},
  author={Li, Wenqi and Milletar{\`\i}, Fausto and Xu, Daguang and Rieke, Nicola and Hancox, Jonny and Zhu, Wentao and Baust, Maximilian and Cheng, Yan and Ourselin, S{\'e}bastien and Cardoso, M Jorge and others},
  booktitle={International workshop on machine learning in medical imaging},
  pages={133--141},
  year={2019},
  organization={Springer}
}

@article{kairouz2021advances,
  title={Advances and open problems in federated learning},
  author={Kairouz, Peter and McMahan, H Brendan},
  journal={Foundations and trends in machine learning},
  volume={14},
  number={1-2},
  pages={1--210},
  year={2021},
  publisher={Emerald Publishing Limited}
}

@article{geiping2020inverting,
  title={Inverting gradients-how easy is it to break privacy in federated learning?},
  author={Geiping, Jonas and Bauermeister, Hartmut and Dr{\"o}ge, Hannah and Moeller, Michael},
  journal={Advances in neural information processing systems},
  volume={33},
  pages={16937--16947},
  year={2020}
}

@article{zhu2019deep,
  title={Deep leakage from gradients},
  author={Zhu, Ligeng and Liu, Zhijian and Han, Song},
  journal={Advances in neural information processing systems},
  volume={32},
  year={2019}
}

@article{clark2003survival,
  title={Survival analysis part I: basic concepts and first analyses},
  author={Clark, Taane G and Bradburn, Michael J and Love, Sharon B and Altman, Douglas G},
  journal={British journal of cancer},
  volume={89},
  number={2},
  pages={232--238},
  year={2003},
  publisher={Nature Publishing Group}
}

@book{kleinbaum1996survival,
  title={Survival analysis a self-learning text},
  author={Kleinbaum, David G and Klein, Mitchel},
  year={1996},
  publisher={Springer}
}

@article{sung2021global,
  title={Global cancer statistics 2020: GLOBOCAN estimates of incidence and mortality worldwide for 36 cancers in 185 countries},
  author={Sung, Hyuna and Ferlay, Jacques and Siegel, Rebecca L and Laversanne, Mathieu and Soerjomataram, Isabelle and Jemal, Ahmedin and Bray, Freddie},
  journal={CA: a cancer journal for clinicians},
  volume={71},
  number={3},
  pages={209--249},
  year={2021},
  publisher={Wiley Online Library}
}

@article{archetti2023scaling,
  title={Scaling survival analysis in healthcare with federated survival forests: A comparative study on heart failure and breast cancer genomics},
  author={Archetti, Alberto and Ieva, Francesca and Matteucci, Matteo},
  journal={Future Generation Computer Systems},
  volume={149},
  pages={343--358},
  year={2023},
  publisher={Elsevier}
}

@article{andreux2020federated,
  title={Federated survival analysis with discrete-time cox models},
  author={Andreux, Mathieu and Manoel, Andre and Menuet, Romuald and Saillard, Charlie and Simpson, Chlo{\'e}},
  journal={arXiv preprint arXiv:2006.08997},
  year={2020}
}

@article{li2025developing,
  title={Developing federated time-to-event scores using heterogeneous real-world survival data},
  author={Li, Siqi and Wang, Ziwen and Shang, Yuqing and Wu, Qiming and Hong, Chuan and Ning, Yilin and Miao, Di and Ong, Marcus Eng Hock and Chakraborty, Bibhas and Liu, Nan},
  journal={Computers in Biology and Medicine},
  volume={197},
  pages={111084},
  year={2025},
  publisher={Elsevier}
}

@article{rahman2022fedpseudo,
  title={Fedpseudo: Pseudo value-based deep learning models for federated survival analysis},
  author={Rahman, Md Mahmudur and Purushotham, Sanjay},
  journal={arXiv preprint arXiv:2207.05247},
  year={2022}
}

@inproceedings{mondrejevski2022flicu,
  title={FLICU: a federated learning workflow for intensive care unit mortality prediction},
  author={Mondrejevski, Lena and Miliou, Ioanna and Montanino, Annaclaudia and Pitts, David and Hollm{\'e}n, Jaakko and Papapetrou, Panagiotis},
  booktitle={2022 IEEE 35th International Symposium on Computer-Based Medical Systems (CBMS)},
  pages={32--37},
  year={2022},
  organization={IEEE}
}

@article{qayyum2022collaborative,
  title={Collaborative federated learning for healthcare: Multi-modal covid-19 diagnosis at the edge},
  author={Qayyum, Adnan and Ahmad, Kashif and Ahsan, Muhammad Ahtazaz and Al-Fuqaha, Ala and Qadir, Junaid},
  journal={IEEE Open Journal of the Computer Society},
  volume={3},
  pages={172--184},
  year={2022},
  publisher={IEEE}
}

@inproceedings{gupta2021hierarchical,
  title={Hierarchical federated learning based anomaly detection using digital twins for smart healthcare},
  author={Gupta, Deepti and Kayode, Olumide and Bhatt, Smriti and Gupta, Maanak and Tosun, Ali Saman},
  booktitle={2021 IEEE 7th international conference on collaboration and internet computing (CIC)},
  pages={16--25},
  year={2021},
  organization={IEEE}
}

@article{chen2023metafed,
  title={Metafed: Federated learning among federations with cyclic knowledge distillation for personalized healthcare},
  author={Chen, Yiqiang and Lu, Wang and Qin, Xin and Wang, Jindong and Xie, Xing},
  journal={IEEE Transactions on Neural Networks and Learning Systems},
  volume={35},
  number={11},
  pages={16671--16682},
  year={2023},
  publisher={IEEE}
}

@article{ogier2022flamby,
  title={Flamby: Datasets and benchmarks for cross-silo federated learning in realistic healthcare settings},
  author={Ogier du Terrail, Jean and Ayed, Samy-Safwan and Cyffers, Edwige and Grimberg, Felix and He, Chaoyang and Loeb, Regis and Mangold, Paul and Marchand, Tanguy and Marfoq, Othmane and Mushtaq, Erum and others},
  journal={Advances in Neural Information Processing Systems},
  volume={35},
  pages={5315--5334},
  year={2022}
}

@article{weinstein2013cancer,
  title={The cancer genome atlas pan-cancer analysis project},
  author={Weinstein, John N and Collisson, Eric A and Mills, Gordon B and Shaw, Kenna R and Ozenberger, Brad A and Ellrott, Kyle and Shmulevich, Ilya and Sander, Chris and Stuart, Joshua M},
  journal={Nature genetics},
  volume={45},
  number={10},
  pages={1113--1120},
  year={2013},
  publisher={Nature Publishing Group}
}

@article{heath2021nci,
  title={The NCI genomic data commons},
  author={Heath, Allison P and Ferretti, Vincent and Agrawal, Stuti and An, Maksim and Angelakos, James C and Arya, Renuka and Bajari, Rosita and Baqar, Bilal and Barnowski, Justin HB and Burt, Jeffrey and others},
  journal={Nature genetics},
  volume={53},
  number={3},
  pages={257--262},
  year={2021},
  publisher={Nature Publishing Group US New York}
}

@article{cox1972regression,
  title={Regression models and life-tables},
  author={Cox, David R},
  journal={Journal of the royal statistical society: Series B (methodological)},
  volume={34},
  number={2},
  pages={187--202},
  year={1972},
  publisher={Wiley Online Library}
}

@inproceedings{lee2018deephit,
  title={Deephit: A deep learning approach to survival analysis with competing risks},
  author={Lee, Changhee and Zame, William and Yoon, Jinsung and Van Der Schaar, Mihaela},
  booktitle={Proceedings of the AAAI conference on artificial intelligence},
  volume={32},
  number={1},
  year={2018}
}

@article{kim2020improved,
  title={Improved survival analysis by learning shared genomic information from pan-cancer data},
  author={Kim, Sunkyu and Kim, Keonwoo and Choe, Junseok and Lee, Inggeol and Kang, Jaewoo},
  journal={Bioinformatics},
  volume={36},
  number={Supplement\_1},
  pages={i389--i398},
  year={2020},
  publisher={Oxford University Press}
}

@article{ishwaran2008random,
    title = {Random survival forests},
    author = {H. Ishwaran and U.B. Kogalur and E.H. Blackstone and M.S.
      Lauer},
    journal = {Ann. Appl. Statist.},
    year = {2008},
    volume = {2},
    number = {3},
    pages = {841--860},
    url = {https://arXiv.org/abs/0811.1645v1},
  }

@article{breiman2001random,
  title={Random forests},
  author={Breiman, Leo},
  journal={Machine learning},
  volume={45},
  number={1},
  pages={5--32},
  year={2001},
  publisher={Springer}
}

@article{liao2024random,
  title={Random survival forest algorithm for risk stratification and survival prediction in gastric neuroendocrine neoplasms},
  author={Liao, Tianbao and Su, Tingting and Lu, Yang and Huang, Lina and Wei, Wei-Yuan and Feng, Lu-Huai},
  journal={Scientific Reports},
  volume={14},
  number={1},
  pages={26969},
  year={2024},
  publisher={Nature Publishing Group UK London}
}

@article{kingma2014adam,
  title={Adam: A method for stochastic optimization},
  author={Kingma, Diederik P and Ba, Jimmy},
  journal={arXiv preprint arXiv:1412.6980},
  year={2014}
}

@article{beutel2020flower,
  title={Flower: A friendly federated learning research framework},
  author={Beutel, Daniel J and Topal, Taner and Mathur, Akhil and Qiu, Xinchi and Fernandez-Marques, Javier and Gao, Yan and Sani, Lorenzo and Li, Kwing Hei and Parcollet, Titouan and de Gusm{\~A}{\c{G}}o, Pedro Porto Buarque and others},
  journal={arXiv preprint arXiv:2007.14390},
  year={2020}
}

\end{document}